\documentclass{article}

\usepackage[preprint]{neurips_2024}

\usepackage[utf8]{inputenc} 
\usepackage[T1]{fontenc}    
\usepackage{hyperref}       
\usepackage{url}            
\usepackage{booktabs}       
\usepackage{amsfonts}       
\usepackage{nicefrac}       
\usepackage{microtype}      
\usepackage{xcolor}

\usepackage{subfigure}

\title{The Local Learning Coefficient: A Singularity-Aware Complexity Measure}
\author{ Edmund Lau\textsuperscript{*}\\
	School of Mathematics and Statistics\\
	University of Melbourne\\
	\texttt{elau1@student.unimelb.edu.au} \\
	\And
        Zach Furman\textsuperscript{*} \\
	Timaeus\\
	\texttt{zach.furman1@gmail.com} \\
	\And
        George Wang \\
	Timaeus\\
	\texttt{george@timaeus.co} \\
	\And
	  Daniel Murfet\\
	School of Mathematics and Statistics\\
	University of Melbourne\\
	\texttt{d.murfet@unimelb.edu.au} \\
        \And
	Susan Wei\\
	Department of Econometrics and Business Statistics\\
	Monash University\\
	\texttt{susan.wei@monash.edu} \\
}

\usepackage{amsmath}
\usepackage{mathtools}
\usepackage{amsthm}
\usepackage{amsfonts}
\usepackage{multirow}
\usepackage{booktabs}
\usepackage{algorithm}
\usepackage{algorithmic}
\usepackage{subcaption}
\usepackage{float}

\newcommand{\E}{\mathrm{E}}

\newcommand{\brac}[1]{\left(#1\right)}

\newcommand{\nbhdLocalFn}{F_n(\nbhd)}

\newcommand{\proxylocalFn}{F_n(w^*,\gamma)}
\newcommand{\proxylocalZn}{Z_n(w^*,\gamma)}
\newcommand{\wbic}{\operatorname{WBIC}}

\newcommand{\nbhd}{B_\gamma(w^*)}

\newcommand{\betastar}{{\beta^*}}
\newcommand{\wstar}{{w^*}}
\newcommand{\hatwstar}{{\hat w_n^*}}
\newcommand{\lambdawstar}{{\lambda(\wstar)}}

\newcommand{\hatlambdawstar}{{\hat \lambda(\wstar)}}
\newcommand{\hatlambdahatwstar}{{\hat \lambda(\hatwstar)}}

\newcommand{\Elocaltempered}{\E_{w|\wstar, \beta, \gamma}}
\newcommand{\ElocaltemperedBetastar}{\E_{w|\wstar,\betastar, \gamma}}
\newcommand{\ElocaltemperedBetastarWstar}{\E_{w|\wstar,\betastar, \gamma}}

\newcommand{\R}{\mathbb{R}}

\DeclarePairedDelimiter{\abs}{\lvert}{\rvert}

\newtheorem{definition}{Definition}
\newtheorem{theorem}{Theorem}

\DeclareMathOperator*{\argmin}{arg\,min}

\begin{document}

\maketitle

\begin{abstract}

The Local Learning Coefficient (LLC) is introduced as a novel complexity measure for deep neural networks (DNNs). Recognizing the limitations of traditional complexity measures, the LLC leverages Singular Learning Theory (SLT), which has long recognized the significance of singularities in the loss landscape geometry. 
This paper provides an extensive exploration of the LLC's theoretical underpinnings, offering both a clear definition and intuitive insights into its application. Moreover, we propose a new scalable estimator for the LLC, which is then effectively applied across diverse architectures including deep linear networks up to 100M parameters, ResNet image models, and transformer language models. Empirical evidence suggests that the LLC provides valuable insights into how training heuristics might influence the effective complexity of DNNs. 
Ultimately, the LLC emerges as a crucial tool for reconciling the apparent contradiction between deep learning's complexity and the principle of parsimony.

\end{abstract}


\section{Introduction}

Occam's razor, a foundational principle in scientific inquiry, suggests that the simplest among competing hypotheses should be selected. This principle has emphasized simplicity and parsimony in scientific thinking for centuries. Yet, the advent of deep neural networks (DNNs), with their multi-million parameter configurations and complex architectures, poses a stark challenge to this time-honored principle. The question arises: how do we reconcile the effectiveness of these intricate models with the pursuit of simplicity?

It is natural to pose this question in terms of model complexity. Unfortunately, many existing definitions of model complexity are problematic for DNNs. For instance, the parameter count, which in classical statistical learning theory is a commonly-used measure of the amount of information captured in a fitted model, is well-known to be inappropriate in deep learning. This is clear from technical results on generalization \citep{zhang2017}, pruning \citep{blalock2020} and distillation \citep{hinton2015}: the amount of information captured in a trained network is in some sense decoupled from the number of parameters. 

As forcefully argued in \citep{weiDeepLearningSingular2022}, the gap in our understanding of DNN model complexity is induced by singularities, hence the need for a singularity-aware complexity measure.
This motivates our appeal to Singular Learning Theory (SLT), which recognizes the necessity for sophisticated tools in studying statistical models exhibiting singularities. Roughly speaking, a model is singular if there are many ways to vary the parameters without changing the function; the more ways to vary without a change, the more singular (and thus more degenerate). DNNs are prime examples of such singular statistical models, characterized by complex degeneracies that make them highly singular. 

SLT continues a longstanding tradition of using \textit{free energy}
as the starting point for measuring model complexity \citep{bialek2001predictability}, and here we see the implications of singularities. Specifically, the free energy, also known as the negative log marginal likelihood, can be shown to asymptotically diverge with sample size $n$ according to the law $an + b \log n + o(\log \log n)$; the coefficient $a$ of the linear term is the minimal loss achievable and the \textbf{coefficient $b$ of the remaining logarithmic divergence} is then taken as the \textbf{model complexity} of the model class. In regular statistical models, the coefficient $b$ is the number of parameters (divided by 2). In singular statistical models, $b$ is not tied to the number of parameters, indicating a different kind of complexity at play.

The LLC arises out of a consideration of the \textit{local} free energy. We explore the mathematical underpinnings of the LLC in Section \ref{section:learning-coefficient}, utilizing intuitive concepts like volume scaling to aid understanding. Our contributions encompass 1) the definition of the new LLC complexity measure, 2) the development of a scalable estimator for the LLC, and 3) empirical validations that underscore the accuracy and practicality of the LLC estimator. In particular, we demonstrate that the estimator is accurate and scalable to modern network size in a setting where theoretical learning coefficients are available for comparison. Furthermore, we show empirically that some common training heuristics effectively control the LLC.

\begin{figure}[b!]
    \centering
    \includegraphics[width=0.99\textwidth, keepaspectratio]{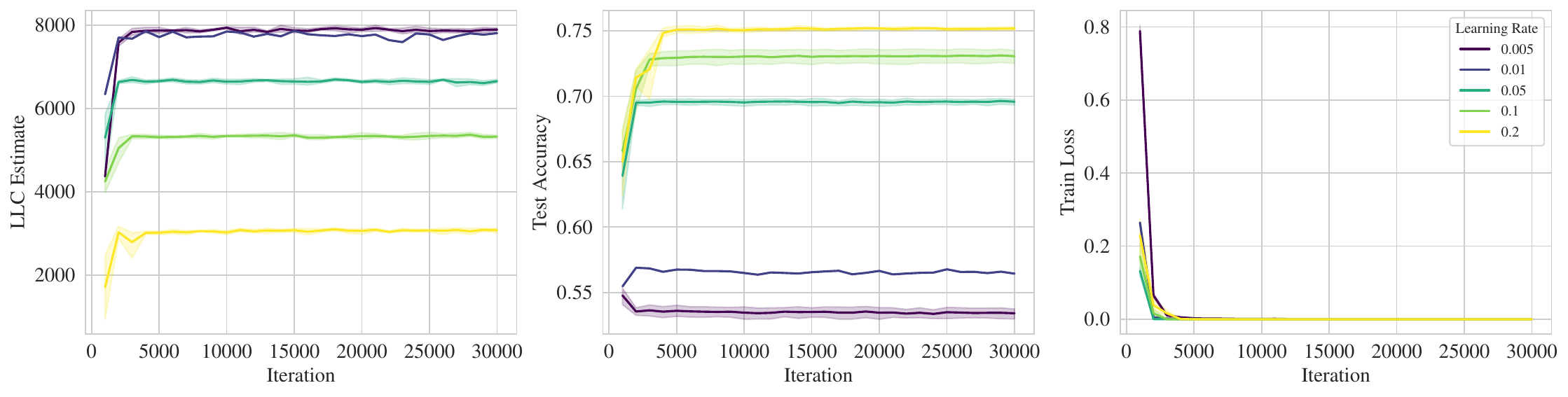} \\
    \includegraphics[width=0.99\textwidth, keepaspectratio]{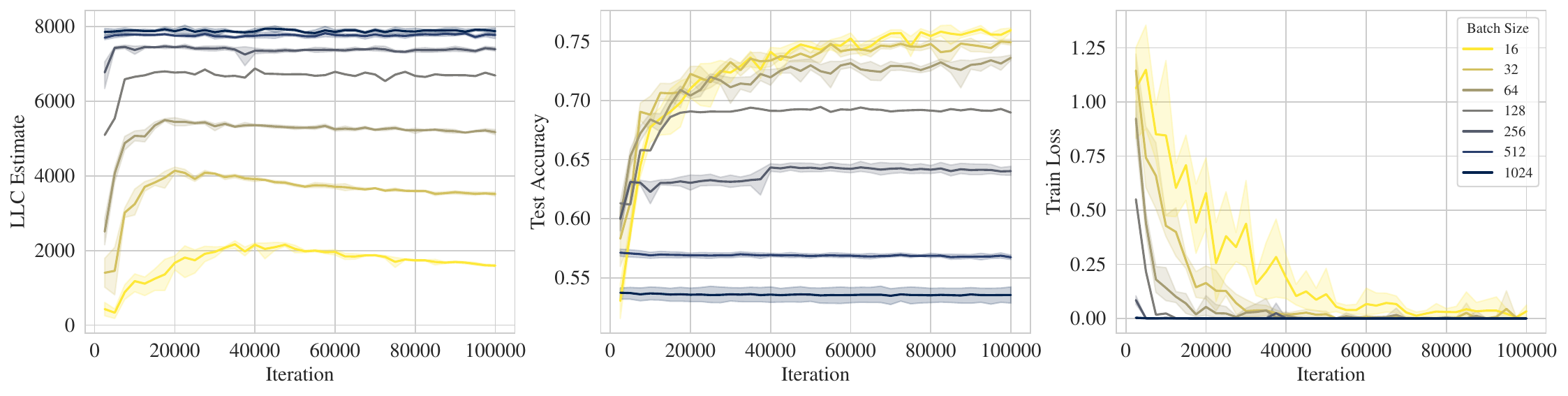}\\
    \includegraphics[width=0.99\textwidth, keepaspectratio]{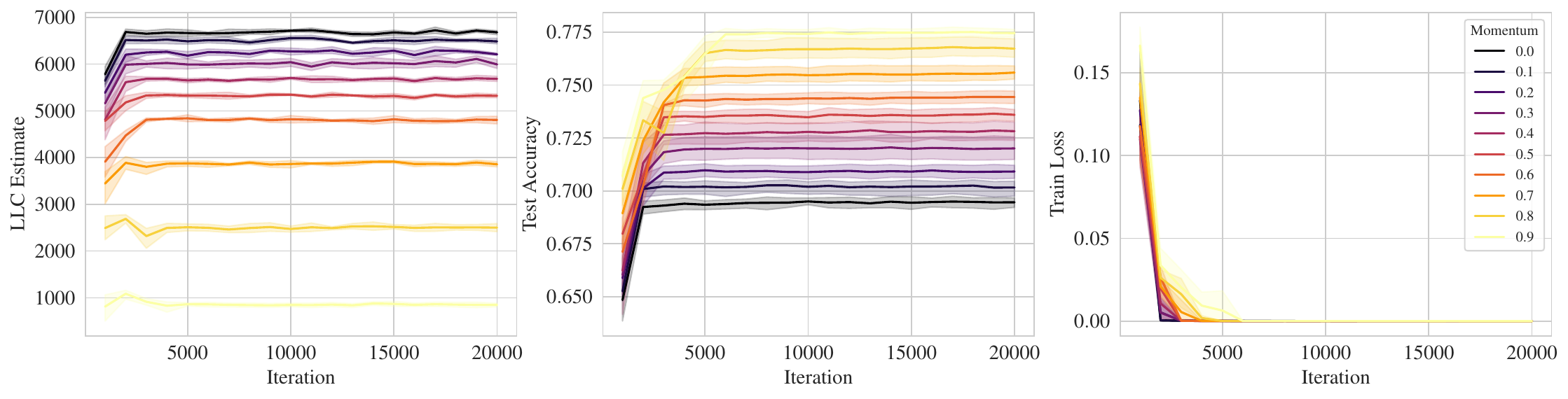}
    \caption{Impact of SGD learning rate (top), batch size (middle) and momentum (bottom) when training ResNet18 on CIFAR10. We plot the LLC estimate (left), test accuracy (middle) and train loss (right) across training time. 
    As the strength of the implicit regularization increases --- through higher learning rate, lower batch size and higher momentum --- LLC decreases (the network gets ``simpler'') and test accuracy increases. 
    Even though most training losses collapse to zero, the LLC can discern the implicit regularization pressure applied by various training heuristics.
    }
    \label{fig:implicit_regularization}
\end{figure}

On this last point, we preview some results; Figure \ref{fig:implicit_regularization} displays the LLC estimate, test accuracy, and and training loss over the course of training ResNet18 on CIFAR10. Loosely speaking, lower LLC
means a less complex, and thus more degenerate, neural network. In each of the rows, lighter colors represent stronger implicit regularization, e.g., higher learning rate, lower batch size, higher SGD momentum, which we see corresponds to a preference for lower LLC, i.e., simpler neural networks. 

\section{Setup}
\label{section:setup}
Let $W \subset \mathbb R^d$ be a compact space of parameters $w \in W$. Consider the model-truth-prior triplet 
\begin{equation}\label{eq:stat_system}
(p(x,y|w), q(x,y), \varphi(w) ),
\end{equation}
where 
$
q(x,y) = q(y|x) q(x)
$
is the true data-generating mechanism,
$
p(x,y|w) = p(y|x,w)q(x)
$
is the posited model with parameter $w$ representing the neural network weights, and $\varphi$ is a prior on $w$. Suppose we are given a training dataset of $n$ input-output pairs, $\mathcal D_n = \{(x_i,y_i)\}_{i=1}^n$, drawn i.i.d. from $q(x,y)$.

To these objects, we can associate the sample negative log likelihood function defined as
$
L_n(w) = - \frac{1}{n} \sum_{i=1}^n \log p(y_i|x_i,w),
$
and its theoretical counterpart defined as
$
L(w) = -\E_{q(x,y)} \log p(y|x,w).
$
It is appropriate to also call $L_n$ and $L$ the training and population loss, respectively, since using the negative log likelihood encompasses many classic loss functions used in machine learning and deep learning, such as mean squared error (MSE) and cross-entropy.

The behavior of the training and population losses is highly nontrivial for neural networks. To properly account for model complexity of neural network models, it is critical to engage with the challenges posed by singularities. To appreciate this, we follow \citep{watanabeAlgebraicGeometryStatistical2009} and make the following distinction between regular and singular statistical models.
A statistical model $p(x, y|w)$ is called \textbf{regular} if it is 1) identifiable, i.e. the parameter to distribution map $w \mapsto p(x, y| w)$ is one-to-one, and 2) its Fisher information matrix $I(w)$ is everywhere positive definite. 
We call a model \textbf{singular} if it is not regular. 
For an introduction to the implications of singular learning theory for deep learning, we refer the readers to Appendix \ref{appendix:slt_background} and further reading in \citep{weiDeepLearningSingular2022}.
We shall assume throughout the triplet \eqref{eq:stat_system} satisfies a few fundamental conditions in SLT \citep{watanabeAlgebraicGeometryStatistical2009}. These conditions are stated and discussed in an accessible manner in \ref{appendix: slt assumptions}.

\section{The local learning coefficient}\label{section:learning-coefficient}




In this paper, we introduce the Local Learning Coefficient (LLC), an extension of \citet{watanabeAlgebraicGeometryStatistical2009}'s global learning coefficient, referred to as simply learning coefficient there.
Below we focus on explaining the LLC through its geometric intuition, specifically as an invariant based on the volume of the loss landscape basin.

For readers interested in the detailed theoretical foundations, we have included comprehensive explanations in the appendices. Appendix A offers a short introduction to the basics of Singular Learning Theory (SLT), and Appendix B sets out formal conditions of the well-definedness of the LLC. This structure ensures that readers with varying levels of familiarity with SLT can engage with the content at their own pace.

\begin{figure*}[ht]
\vskip 0.2in
\begin{center}
  \subfigure{
    \includegraphics[width=\textwidth]{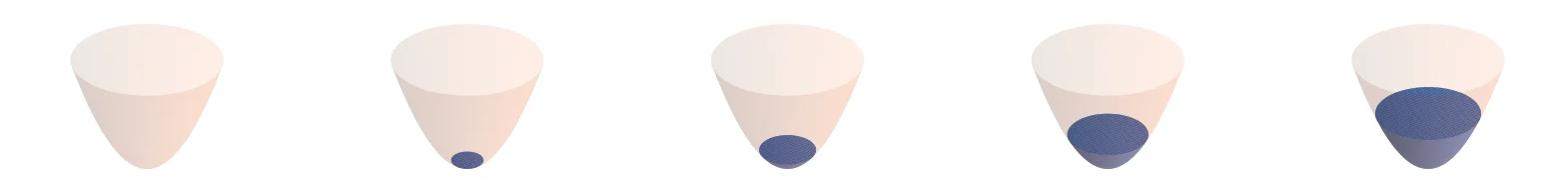}
    \label{fig:regular_volume_scaling}
  }
  \subfigure{
    \includegraphics[width=\textwidth]{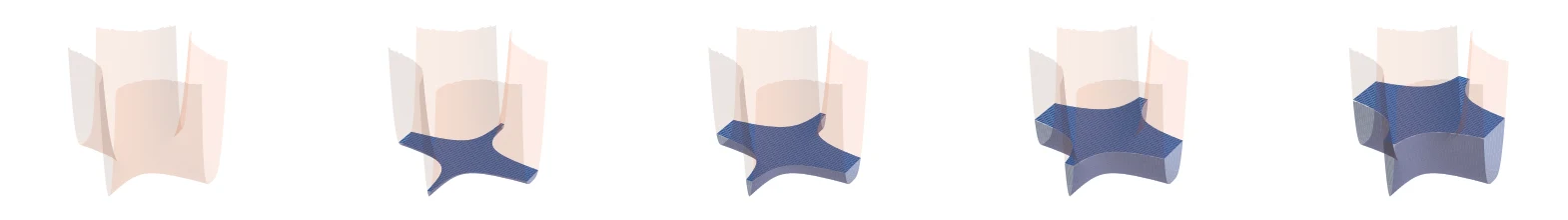}
    \label{fig:singular_volume_scaling}
  }
  \caption{The LLC $\lambdawstar$ measures \textit{volume scaling} around a local minimum of the loss. Top: the area (or ``2D-volume'') under the loss $L(w_1, w_2) = {w_1}^2 + {w_2}^2$ scales as $V(\epsilon) \propto \epsilon$ around $\wstar = (0,0)$, so $\lambdawstar = 1$. Bottom: the area under the loss $L(w_1, w_2) = {w_1}^2 {w_2}^4$ scales as $V(\epsilon) \propto \epsilon^{1/4}$ around $\wstar = (0,0)$, so $\lambdawstar = \frac{1}{4}$. Lower $\lambdawstar$ means more volume: as we approach the minimum ($\epsilon \rightarrow 0$), there is much more volume available for the bottom loss function compared to the top for any given small tolerance $\epsilon$. Reproduced with permission from \citep{hoogland2023}.}
  \label{fig:volume_scaling}
\end{center}
\vskip -0.2in
\end{figure*}

\subsection{Complexity via counting low loss parameters}\label{section:complexity_counting}


At a local minimum of the population loss landscape there is a natural notion of complexity, given by the number of bits required to specify the minimum to within a tolerance $\epsilon$. This idea is well-known in the literature on minimum description length \citep{grunwald2019minimum} and was used by \citep{hochreiter1997} in an attempt to quantify the complexity of a trained neural network. However, a correct treatment has to take into consideration the degeneracy of the geometry of the population loss $L$, as we now explain.

Consider a local minimum $w^*$ of the population loss $L$ and a closed ball $B(w^*)$ centered on $w^*$ such that for all $w \in B(w^*)$ we have $L(w) \ge L(w^*)$. Given a tolerance $\epsilon > 0$ we can consider the set of parameters $B(w^*, \epsilon) = \{w \in B(w^*) \mid L(w) - L(w^*) < \epsilon\}$ whose loss is within the tolerance, the volume of which we define to be
\begin{equation}
    V(\epsilon) := \operatorname{Vol}( B(w^*, \epsilon) ) = \int_{B(w^*, \epsilon)} dw\,.
\end{equation}
The minimal number of bits to specify this set within the ball is
\begin{equation}\label{eq:number_of_bits}
    - \log_2( V(\epsilon) / \operatorname{Vol}(B(w^*)) )\,.
\end{equation}
This can be taken as a measure of the complexity of the set of low loss parameters near $w^*$. However, as it stands this notion depends on $\epsilon$ and has no intrinsic meaning. Classically, this is addressed as follows: if a model is regular and thus $L(w)$ is locally quadratic around $w^*$, the volume satisfies a law of the form
$
V(\epsilon) \approx c\epsilon^{d/2}, 
$
where $c$ is a constant that depends on the curvature of the basin around the local minimum $w^*$, and $d$ is the dimension of $W$. 

This explains why $\frac{d}{2}$ is a valid measure of complexity in the regular case, albeit one that cannot distinguish $w^*$ from any other local minima. The curvature $c$ is less significant than the scaling exponent, but \emph{can} distinguish local minima and $\log c$ is sometimes used as a complexity measure.

The population loss of a neural network is \emph{not} locally quadratic near its local minima, from which it follows that the functional form of the volume $V(\epsilon)$ is more complicated than in the regular case. The correct functional form was discovered by \citet{watanabeAlgebraicGeometryStatistical2009} and we adapt it here to a local neighborhood of a parameter (see 
Appendix \ref{appendix:slt_background} for details):

\begin{definition}[The Local Learning Coefficient (LLC), $\lambdawstar$]\label{defn:llc} 
 There exists a unique rational\footnote{The fact that $\lambdawstar$ is \textit{rational-valued}, and not real-valued, as one would naively assume, is a deep fact of algebraic geometry derived from the celebrated Hironaka's resolution of singularities.} number $\lambdawstar$, a positive integer $m(\wstar)$ and some constant $c > 0$ such that asymptotically as $\epsilon \to 0$, 
\begin{equation}
 V(\epsilon) = c \epsilon^{\lambdawstar} (-\log \epsilon)^{m(w^*)-1} + o(\epsilon^{\lambdawstar} (-\log \epsilon)^{m(w^*)-1}). \label{eq:volume scaling}
\end{equation}
We call $\lambdawstar$ the Local Learning Coefficient (LLC), and $m(w^*)$ the local multiplicity.
\end{definition}

In the case where $m(\wstar)=1$, the formula simplifies, and
\begin{equation}
    V(\epsilon) \propto \epsilon^{\lambdawstar} \label{eq: simplified volume scaling}
\end{equation}
Thus, the LLC $\lambdawstar$ is the (asymptotic) \textit{volume scaling exponent} near a minimum $\wstar$ in the loss landscape: increasing the error tolerance by a factor of $a$ increases the volume by a factor of $a^{\lambdawstar}$. Applying Equation \eqref{eq:number_of_bits}, the number of bits needed to specify $V(\epsilon)$ within $B(w^*)$, for sufficiently small $\epsilon$ and in the case $m(w^*) = 1$, is approximated by
\begin{equation*}
-\lambdawstar \log_2 \epsilon + O(\log_2 \log_2 \epsilon)\,.
\end{equation*}
Informally, the LLC tells us the number of additional bits needed to halve an already small error of $\epsilon$: 
\begin{equation*}
    -\log_2\Big[ V(\tfrac{\epsilon}{2}) / V(\epsilon) \Big] \approx -\lambdawstar \log_2 \frac{\epsilon}{2} + \lambdawstar \log_2 \epsilon  = \lambdawstar
\end{equation*}
See Figure \ref{fig:volume_scaling} for simple examples of the LLC as a scaling exponent. We note that, in contrast to the regular case, the scaling exponent $\lambdawstar$ depends on the underlying data distribution $q(x,y)$. 


The (global) learning coefficient $\lambda$ was defined by \citet{watanabe2001}. If $w^*$ is taken to be a global minimum of the population loss $L(w)$, and the ball $B(w^*)$ is taken to be the entire parameter space $W$, then we obtain the global learning coefficient $\lambda$ as the scaling exponent of the volume $V(\epsilon)$. The learning coefficient and related quantities like the WBIC \citep{watanabeWidelyApplicableBayesian2013} have historically seen significant application in Bayesian model selection (e.g. \citealp{endo2020, fontanesi2019, hooten2015, sharma2017, kafashan2021, semenova2020}).

In Appendix \ref{appendix:reparam_invariance}, we prove that the LLC is invariant to \textit{local diffeomorphism}: that is, roughly, a locally smooth and invertible change of variables. This property is motivated by the desire that a good complexity measure should not be confounded by superficial differences in how a model is represented: two models which are essentially the same should have the same complexity.

\section{LLC estimation}
\label{section:estimate_lambda}
Having established the intuition behind the theoretical LLC in terms of volume scaling, we now turn to the task of estimating it. As described in the introduction, the LLC is a coefficient in the asymptotic expansion of the local free energy. It is this fact that we leverage for estimation. There is a mathematically rigorous link between volume scaling and the appearance of the LLC in the local free energy which we will not discuss in detail; see \citep[Theorem 7.1]{watanabeAlgebraicGeometryStatistical2009}.

We first introduce what we call the idealized LLC estimator which is theoretically sound, albeit intractable from a computational point of view. In the sections that follow the introduction of the idealized LLC estimator, we walk through the steps taken to engineer a practically implementable version of the idealized LLC estimator.

\subsection{Idealized LLC estimator}
Consider the following integral 
\begin{equation}
    Z_n(\nbhd) = \int_{\nbhd} \exp\{-n L_n(w)\} \varphi(w) \,dw,
    \label{eq:localZn_nbhd}
\end{equation}
where $L_n(w)$ is again the sample negative log likelihood, $\nbhd$ denotes a small ball of radius $\gamma$ around $\wstar$ and $\varphi$ is a prior over model parameters $w$. If \eqref{eq:localZn_nbhd} is high, there is high posterior concentration around $\wstar$. In this sense, \eqref{eq:localZn_nbhd} is a measure of the concentration of low-loss solutions near $\wstar$.
Next consider a log transformation of $Z_n(\nbhd)$ with a negative sign, i.e.,
\begin{equation}
    \nbhdLocalFn = - \log Z_n(\nbhd).
    \label{eq:localFn_nbhd}
\end{equation}
This quantity is sometimes called (negative) log marginal likelihood or free energy, depending on the discipline. Given a local minimum $\wstar$  of the population negative log likelihood $L(w)$, it can be shown using the machinery of SLT that, asymptotically in $n$, we have
\begin{equation}
\nbhdLocalFn = n \underbrace{L_n(\wstar)}_{\text{energy}} + \underbrace{\lambda(\wstar)}_{\text{entropy}} \log n + o_p(\log \log n), 
\label{eq:localFn_nbhd_energy_entropy}
\end{equation}
where $\lambdawstar$ is the theoretical LLC expounded in Section \ref{section:learning-coefficient}. Remarkably, the asymptotic approximation in \eqref{eq:localFn_nbhd_energy_entropy} holds even for singular models such as neural networks; further discussion on this can be found in Appendix \ref{appendix:well_definedness}.

The asymptotic approximation in \eqref{eq:localFn_nbhd_energy_entropy} suggests that a reasonable estimator of $\lambdawstar$ might come from re-arranging \eqref{eq:localFn_nbhd_energy_entropy} to give what we call the idealized LLC estimator,
\begin{equation}
    \hat \lambda^{\mathrm{idealized}}(\wstar) = \frac{\nbhdLocalFn-n L_n(\wstar)}{\log n}.
    \label{eq:idealized_lambda}
\end{equation}
But as indicated by the name, the idealized LLC estimator cannot be easily implemented; computing or even MCMC sampling from the posterior to estimate $\nbhdLocalFn$ is made no less challenging 
by the need to confine sampling to the neighborhood $\nbhd$. In what follows, we use the idealized LLC estimator as inspiration for a practically-minded LLC estimator.

\subsection{Surrogate for enforcing $\nbhd$}
The first step towards a practically-minded LLC estimator is to circumvent the constraint posed by the neighborhood $\nbhd$. To this end, we introduce  a localizing Gaussian prior that acts as a \textit{surrogate for enforcing the domain of integration} given by $B_\gamma(\wstar)$. Specifically, let 
\begin{equation*}
    \varphi_\gamma(w) \propto \exp\left \{ -\frac{\gamma}{2} ||w||^2_2 \right\}
\end{equation*}
be a Gaussian prior centered at the origin with scale parameter $\gamma>0$. We replace \eqref{eq:localZn_nbhd} with
$$
\proxylocalZn = \int \exp\{-n L_n(w)\} \varphi_\gamma(w-w^*) \,dw,
$$ 
which, for $\beta=1$, can also be recognized as the normalizing constant to the posterior distribution given by
\begin{equation}
p(w|w^*,\beta, \gamma) \propto \exp \left \{-n \beta L_n(w)-\frac{\gamma}{2}||w-w^*||_2^2 \right \},
\label{eq:local_tempered_posterior}
\end{equation}
where $\beta>0$ plays the role of an inverse temperature.  Large values of $\gamma$ force the posterior distribution in \eqref{eq:local_tempered_posterior} to stay close to $\wstar$. A word on the notation $p(w|w^*,\beta, \gamma)$: this is a distribution in $w$ solely, the parameters $\wstar, \beta, \gamma$ are fixed, hence the normalizing constant to \eqref{eq:local_tempered_posterior} is an integral over $w$ only. 
As $\proxylocalZn$ is to be viewed as a proxy to \eqref{eq:localZn_nbhd}, we shall accordingly treat 
\begin{equation*}
\proxylocalFn \coloneqq -\log \proxylocalZn
\end{equation*}
as a proxy for $\nbhdLocalFn$ in \eqref{eq:localFn_nbhd}. Although it is tempting at this stage to simply drop $\proxylocalFn$ into the idealized LLC estimator in place of $F_n(B_\gamma(\wstar))$, we have to address estimation of $\proxylocalFn$, the subject of the next section. 

\subsection{The LLC estimator}
\label{sec:llc_estimator}
Let us denote the expectation of a function $f(w)$ with respect to the posterior distribution in \eqref{eq:local_tempered_posterior} as 
\begin{equation*}
\Elocaltempered f(w) \coloneqq \int f(w) p(w|w^*,\beta, \gamma) \,dw.
\end{equation*}
Consider the quantity
\begin{equation}
\ElocaltemperedBetastarWstar [nL_n(w)]
    \label{eq:localWBIC}
\end{equation}
where the inverse temperature is deliberately set to $\beta^* = 1 / \log n$. The quantity in \eqref{eq:localWBIC} may be regarded as a localized version of the widely applicable Bayesian information criterion (WBIC) first introduced in \citep{watanabeWidelyApplicableBayesian2013}. It can be shown that \eqref{eq:localWBIC} is a good estimator of $\proxylocalFn$ in the following sense: the leading order terms of \eqref{eq:localWBIC} match those of $\proxylocalFn$ when we perform an asymptotic expansion in sample size $n$. This justifies using \eqref{eq:localWBIC} to estimate $\proxylocalFn$. Further discussion on this can be found in in Appendix \ref{appendix:consistency_local_WBIC}

Going back to \eqref{eq:idealized_lambda}, we approximate $\nbhdLocalFn$ first with $\proxylocalFn$, which is further estimated by \eqref{eq:localWBIC}. We are finally ready to define the LLC estimator. 
\begin{definition}[Local Learning Coefficient (LLC) estimator]
Let $\wstar$ be a local minimum of $L(w)$. Let $\betastar=1/\log n$.
The associated local learning coefficient estimator is given by
\begin{equation}
    \hat \lambda(w^*) \coloneqq n\betastar \left [\ElocaltemperedBetastar L_n(w)  - L_n(\wstar) \right ].
    \label{eq:hatlambda}
\end{equation}
\end{definition}
Note that $\hat \lambda(w^*)$ depends on $\gamma$ but we have suppressed this in the notation.
Let us ponder the pleasingly simple form that is the LLC estimator. The expectation term in \eqref{eq:hatlambda} is a measure of the loss $L_n$ under perturbation near $\wstar$. 
If the perturbed loss, under this expectation, is very close to $L_n(\wstar)$, then $\hatlambdawstar$ is small. This accords with our intuition that if $\wstar$ is simple, its loss should not change too much under reasonable perturbations. 
Finally we note that in applications, we use the empirical loss $L_n(w)$ to determine a critical point of interest, i.e., 
$
\hatwstar \coloneqq \argmin_{w} L_n(w)\,.
$
We lose something by plugging in $\hatwstar$ to \eqref{eq:hatlambda} directly since we end up using the dataset $\mathcal D_n$ twice. However we do not observe adverse effects in our experiments, see Figure \ref{fig:scalability} for an example. 

\subsection{The SGLD-based LLC estimator}\label{section:SGLD_based_wbic}
The LLC estimator defined in \eqref{eq:hatlambda} is not prescriptive as to how the expectation with respect to the posterior distribution should actually be approximated. A wide array of MCMC techniques are possible. However, to be able to estimate the LLC at the scale of modern deep learning, we must look at efficiency. In practice, \textit{the computational bottleneck to implementing \eqref{eq:hatlambda} is the MCMC sampler}. In particular, traditional MCMC samplers must compute log-likelihood gradients across the \textit{entire} training dataset, which is prohibitively expensive at modern dataset sizes.


If one modifies these samplers to take minibatch gradients instead of full-batch gradients, this results in \textit{stochastic-gradient MCMC}, the prototypical example of which is \citep{wellingBayesianLearningStochastic2011}'s Stochastic Gradient Langevin Dynamics (SGLD). The computational cost of this sampler is much lower: roughly the cost of a single SGD step times the number of samples required.
The standard SGLD update applied to sampling \eqref{eq:local_tempered_posterior} at the optimal temperature $\beta^*$ required for the LLC estimator is given by
\begin{align*}
\Delta w_t &= \frac{\epsilon}{2} \left ( \frac{\beta^* n}{m} \sum_{(x, y) \in B_t} \nabla \log p(y | x, w_t) + \gamma (w^*-w_t) \right )+ N(0, \epsilon)
\end{align*}
where $B_t = \{(x_{i},y_{i})\}_{i=1}^m$ is a randomly sampled minibatch of samples of size $m$ for step $t$ and $\epsilon$ controls both step size and variance of injected Gaussian noise. Crucially, the log-likelihood gradient is evaluated using mini-batches. In practice, we choose to shuffle the dataset once and partition it into a sequence of size $m$ segments as minibatches instead of drawing fresh random samples of size $m$. 

Let us now suppose we have obtained $T$ approximate samples $\{w_1, w_2, \dots, w_T\}$ of the tempered posterior distribution at inverse temperature $\beta^*$ via SGLD. This is usually taken from the SGLD trajectory after burn-in. We can then form what we call the \textbf{SGLD-based LLC estimator}, 
\begin{equation}
\hat \lambda^{\mathrm{SGLD}}(\wstar) := n\betastar \left [ \frac{1}{T} \sum_{t=1}^T L_n(w_t) - L_n(\wstar) \right ].    
\end{equation}

For further computation saving, we also recycle the forward passes that compute $L_m(w_t)$, which is required for computing $\nabla_w L_m(w_t)$ via back-propagation, as unbiased estimate of $L_n(w_t)$. Here by $L_m(w_t)$, we mean $-\frac{1}{m} \sum_{(x,y) \in B_t} \log p(y|x,w_t)$, though the notation suppresses the dependence on $B_t$ for brevity. Pseudocode for this minibatch version of the SGLD-based LLC estimator is provided in Appendix \ref{appendix:minibatch_hatlambdawstar}. Henceforth, when we say the SGLD-based LLC estimator, we are referring to the minibatch version. In Appendix \ref{appendix:recommendations}, we give a comprehensive guide on best practices for implementing the SGLD-based LLC estimator including choices for $\gamma, \epsilon$, number of SGLD iterations, and required burn-in.

\section{Experiments}\label{section:experiments}


The goal of our experiments is to give evidence that the LLC estimator is accurate, scalable and can reveal insights on deep learning practice. Throughout the experiments, we implement the minibatch version of the SGLD-based LLC estimator presented in the pseudo-algorithm in Appendix \ref{appendix:minibatch_hatlambdawstar}.

There are also a number of experiments we performed that are relegated to the appendices due to space constraints. They include 
\begin{itemize}
    \item deploying LLC estimation on transformer language models (Section \ref{appendix:expt_language})
    \item an experiment on a small ReLU network verifying that SGLD sampler is just as accurate as one that uses full-batch gradients (Appendix \ref{appendix:expt_mala_vs_sgld})
    \item an experiment comparing the optimizers SGD and entropy-SGD with the latter having intimate connection to our notion of complexity (Appendix \ref{appendix:expt_esgd})
    \item  an experiment verifying the scaling invariance property (Appendix \ref{appendix:expt_rescaling}) set out in Appendix \ref{appendix:reparam_invariance}
\end{itemize}
Every experiment described in the main text below has an accompanying section in the Appendix that offers full experimental details, further discussion, and possible additional figures/tables.

\subsection{LLC for Deep Linear Networks (DLNs)} \label{section:expt_dln}

In this section, we verify the \textit{accuracy} and \textit{scalability} of our LLC estimator against theoretical LLC values in deep linear networks (DLNs) up to 100M parameters. Recall DLNs are fully-connected feedforward neural networks with identity activation function. The input-output behavior of a DLN is obviously trivial; it is equivalent to a single-layer linear network obtained by multiplying together the weight matrices. However, the \textit{geometry} of such a model is highly non-trivial --- in particular, the optimization dynamics and inductive biases of such networks have seen significant research interest \citep{saxe2013, ji2018, arora2018}.

Thus one reason we chose to study DLN model complexity is because they have long served as an important sandbox for deep learning theory. Another key factor is the recent clarification of theoretical LLCs in DLNs by \citet{aoyagi2024}, making DLNs the most realistic setting where theoretical learning coefficients are available. The significance of \citet{aoyagi2024} lies in the substantial technical difficulty of deriving theoretical global learning coefficients (and, by extension, theoretical LLCs) which means that these coefficients are generally unavailable except in a few exceptional cases, with most of the research conducted decades ago \citep{yamazaki2005a, aoyagi2005, yamazaki2003, yamazaki2005b}. Further details and discussion of the theoretical result in \citet{aoyagi2024} can be found in Appendix \ref{appendix: aoyagi dln lambda}.

\begin{figure}[ht!]
\begin{center}
\includegraphics[width=0.49\columnwidth]{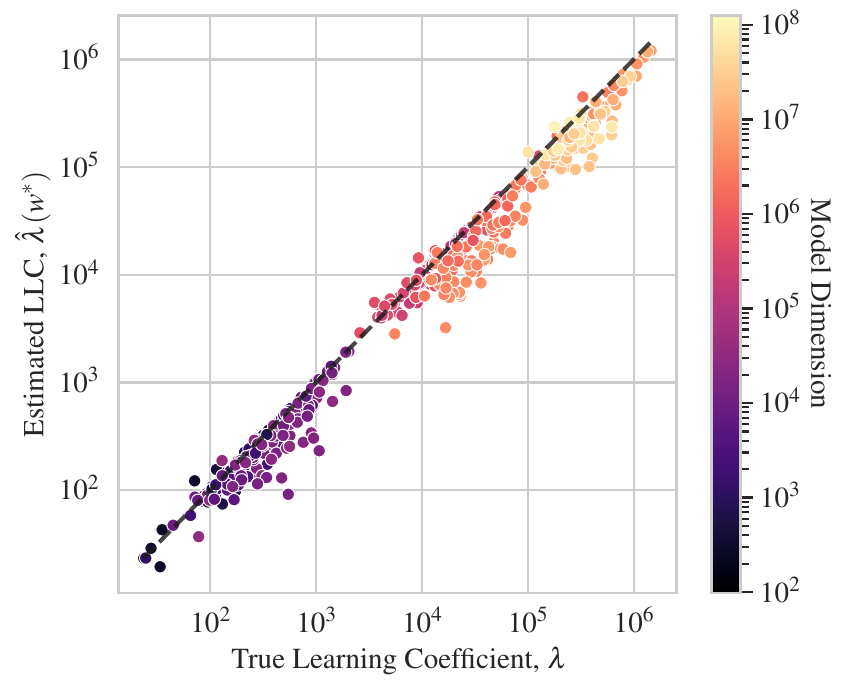}
\includegraphics[width=0.49\columnwidth]{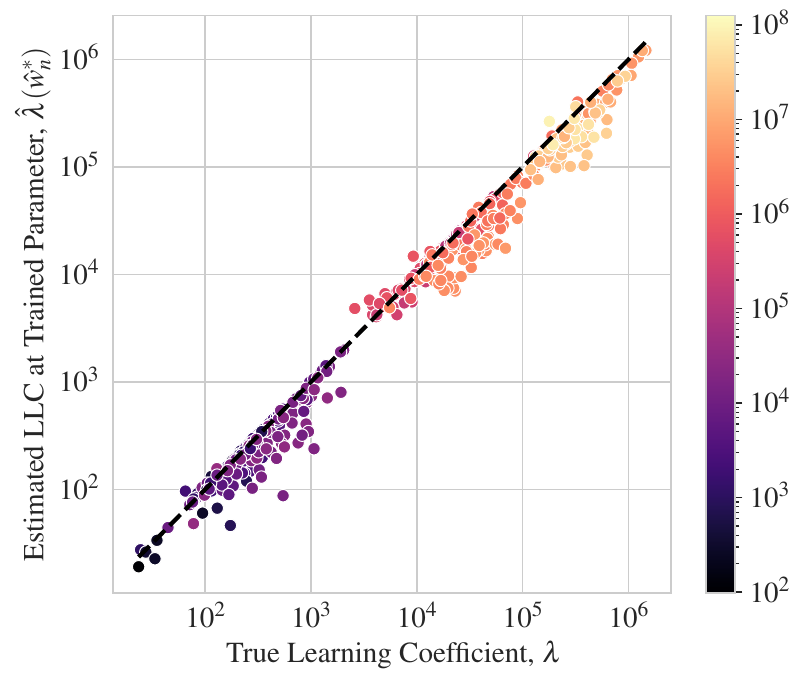}
\caption{Estimated LLC against true learning coefficient; model dimension shown in color. On the left, we evaluate the LLC estimator at a global minimum, $\wstar$, of the population loss. On  the right, we evaluate the LLC estimator at a minimum, $\hat w_n^*$, found by SGD. Fortunately, we do not see an adverse effect of using the training data twice, a minor concern we had raised at the end of Section \ref{sec:llc_estimator}. The estimated LLCs accurately measures the learning coefficient $\lambda$ up to 100 million parameters in deep linear networks, as compared to known theoretical values (dashed line).  See Figure \ref{fig:lambdahat-vs-lambda linear scale} for linear-scale plots.}
\label{fig:scalability}
\end{center}
\end{figure}


The results are summarized in Figure \ref{fig:scalability}. Our LLC estimator is able to accurately estimate the learning coefficient in DLNs up to 100M parameters. We further show that accuracy is maintained even if (as is typical in practice) one does not evaluate the LLC at a local minimum of the population loss, but is instead forced to use SGD to first find a local minimum $\hat w_n^*$. Further experimental details and results for this section are detailed in Appendix \ref{appendix:expt_dln}. Overall it is quite remarkable that our LLC estimator, after the series of engineering steps described in Section \ref{section:estimate_lambda}, maintains such high levels of accuracy.

\subsection{LLC for ResNet}\label{section:expt_cifar10}
Here, we empirically test whether implicit regularization effectively induces a preference for simplicity as manifested by a \textit{lower} LLC. We examine the effects of SGD learning rate, batch size, and momentum on the LLC. Note that we do so in isolation, i.e., we do not look at interactions between these factors. For instance, when we vary the learning rate, we hold everything else constant including batch size and momentum values. 

We perform experiments on ResNet18 trained on CIFAR10 and show the results in Figure \ref{fig:implicit_regularization}. We see that the training loss reaches zero in most instances and therefore on its own cannot distinguish between the effect of the implicit regularization. In contrast, we see that there is a consistent pattern of ``stronger implicit regularization = higher test accuracy = lower LLC". Specifically, higher learning rate, lower batch size, higher momentum all apply stronger implicit regularization which is reflected in lower LLC. Full experimental details can be found in Appendix \ref{appendix:expt_cifar}. Note that in Figures \ref{fig:implicit_regularization} we employed SGD \textit{without} momentum in the top two rows. We repeat the experiments in these top two rows for SGD \textit{with} momentum; the associated results in Figure \ref{fig:llc_curve_implicit_reg_appendix} support very similar conclusions. We also conducted some \textit{explicit} regularization experiments involving L2 regularization (Figure \ref{fig:llc_curve_varyl2reg_appendix} in Appendix \ref{appendix:expt_cifar}) and again conclude that stronger regularization is accompanied by lower LLC. 

In contrast to the LLC estimates for DLN, the LLC estimates for ResNet18 cannot be calibrated as we do not know the true LLC values. With many models in practice, we find value in the LLC estimates by comparing their relative values between models with shared context. For instance, when comparing LLC values we might hold everything constant while varying one factor such as SGD batch size, learning rate, or momentum as we have done in this section.

\section{Related work}
\label{sec:related}

We briefly cover related work here; more detail may be found in Appendix \ref{appendix: related work}. The primary reference for the theoretical foundation of this work, known as SLT, is \citet{watanabeAlgebraicGeometryStatistical2009}. The global learning coefficient, first introduced by \citep{watanabe2001}, provides the asymptotic expansion of the free energy, which is equivalent to the negative log Bayes marginal likelihood, an all-important important quantity in Bayesian analysis. Subsequent research has utilized algebraic-geometric tools to calculate the learning coefficient for various machine learning models \citep{yamazaki2005a, aoyagi2005, aoyagi2024, yamazaki2003, yamazaki2005b}.  SLT has also enhanced the understanding of model selection criteria in Bayesian statistics. Of particular relevance to this work is \citet{watanabeWidelyApplicableBayesian2013}, which introduced the WBIC estimator of free energy. This estimator has been applied in various practical settings \citep{endo2020, fontanesi2019, hooten2015, sharma2017, kafashan2021, semenova2020}.

The LLC can be seen as a singularity-aware version of basin broadness measures, which attempt to connect geometric "broadness" or "flatness" with model complexity \citep{hochreiter1997, jiangFantasticGeneralizationMeasures2019}. In particular, the LLC \textit{estimator} bears resemblance to PAC-Bayes inspired flatness/sharpness measures \citep{neyshabur2017}, but takes into account the non-Gaussian nature of the posterior distribution in singular models.

The LLC is set apart from classic model complexity measures such as Rademacher complexity \citep{koltchinskii2000} and the VC dimension \citep{vapnik1971} because it measures the complexity of a specific model $p(y|x,w)$ rather than the complexity over the function class $\{f(x|w): w\}$ where $f$ is the neural network function. This makes the LLC an appealing tool for understanding the interplay between function class, data properties, and training heuristics.

Concurrent work by \citep{chenLearningCapacityMeasure2023} proposes a measure called the learning capacity, which can be viewed as a finite-$n$ version of the learning coefficient, and investigates its behavior as a function of training size $n$.

\section{Outlook}


An exciting direction of future research is to study the role of the LLC in detecting phase transitions and emergent abilities in deep learning models. A first step in this direction was undertaken in \citep{chen2023dynamical} where the energy (training loss) and entropy (both estimated and theoretical LLC) were tracked
as training progresses; it was observed that energy and entropy proceed along staircases in opposing directions. Further, \citet{hoogland2024} showed how the estimated LLC can be used to detect phase transitions in the formation of in-context learning in transformer language models. It is natural to wonder if the LLC could shed light on the hypothesis that SGD-trained neural networks sequentially learn the target function with a ``saddle-to-saddle" dynamic. Previous theoretical works had to devise complexity measures on a case-by-case basis \citep{abbe23a, berthier2023incremental}. We posit that the free energy perspective could offer a more unified and general approach to understanding the intricate dynamics of learning in deep neural networks by accounting for competition between model fit and model complexity during training.

\bibliographystyle{apalike}
\bibliography{references}

\newpage
\appendix
\counterwithin{figure}{section}

\section{Background on Singular Learning Theory }\label{appendix:slt_background}

Most models in machine learning are singular: they contain parameters where the Fisher information matrix is singular. While these parameters where the Fisher information is degenerate form a measure zero subset, their effect is far from negligible. Singular Learning Theory (SLT) shows that the geometry in the neighbourhood of these degenerate points determines the asymptotics of learning \citep{watanabeAlgebraicGeometryStatistical2009}. The theory explains observable effects of degeneracy in common machine learning models under practical settings \citep{watanabe2018} and has the potential to account for important phenomena in deep learning \citep{weiDeepLearningSingular2022}. 
The central quantity of SLT is the learning coefficient $\lambda$. Many notable SLT results are conveyed through the learning coefficient which can be thought of as the complexity of the model class relative to the true distribution. In this section we carefully define the (global) learning coefficient $\lambda$, which we then contrast with the \emph{local} learning coefficient $\lambda(w^*)$.


For notational simplicity, we consider here the unsupervised setting where we have the model $p(x|w)$ parameterised by a compact parameter space $W \subset \R^d$. We assume fundamental conditions I and II of \citep[\S 6.1, \S 6.2]{watanabeAlgebraicGeometryStatistical2009}. In particular $W$ is defined by a finite set of real analytic inequalities, at every parameter $w \in W$ the distribution $p(x |w)$ should have the same support as the true density $q(x)$, and the prior density $\varphi(w) = \varphi_1(w) \varphi_2(w)$ is a product of a positive smooth function $\varphi_1(w)$ and non-negative real analytic $\varphi_2(w)$. We refer to
\begin{equation}\label{eq:unsupervised_triplet}
(p(x|w), q(x), \varphi(w) )
\end{equation}
as the model-truth-prior triplet.

Let $K(w)$ be the Kullback-Leibler divergence between the truth and the model
\begin{equation*}
K(w) \coloneqq \operatorname{KL}\big(q(x) \,\Vert\, p(x | w) \big) = \int q(x) \log\Big[ q(x) / p(x|w) \Big] dx
\end{equation*}
and define the (average) negative log likelihood to be
\begin{equation*}
L(w) \coloneqq - \int q(x) \log p(x|w) dx = K(w) - S
\end{equation*}
where $S$ is the entropy of the true distribution. Set $K_0 = \inf_{w \in W} K(w)$. Let 
\begin{equation*}
W_0 \coloneqq \{w \in W: K(w) = K_0\}
\end{equation*}
be the set of \textbf{optimal parameters}. We say the \textbf{truth is realizable by the model} if $K_0 = 0$. We do not assume that our models are regular or that the truth is realizable, but we do assume the model-truth-prior triplet satisfies the more general condition of \textbf{relative finite variance} \citep{watanabeWidelyApplicableBayesian2013}. We also assume that there exists $w_0^*$ in the interior of $W$ satisfying $K(w_0^*) = K_0$.

Following \citet{watanabeAlgebraicGeometryStatistical2009} we define:

\begin{definition}\label{defn:zeta_function}
    The \textbf{zeta function} of \eqref{eq:unsupervised_triplet} is defined for $\operatorname{Re}(z) > 0$ by
    \[
    \zeta(z) = \int_W \left(K(w) - K_0\right)^z \varphi(w) dw
    \]
    and can be analytically continued to a meromorphic function on the complex plane with poles that are all real, negative and rational \citep[Theorem 6.6]{watanabeAlgebraicGeometryStatistical2009}. Let $-\lambda \in \mathbb R$ be the largest pole of $\zeta$ and $m$ its multiplicity. Then, the \textbf{learning coefficient} and its \textbf{multiplicity} of the triple \eqref{eq:unsupervised_triplet} are defined to be $\lambda$ and $m$ respectively. 
\end{definition}

When $p(x|w)$ is a singular model, $W_0$ is an analytic variety which is in general positive dimensional (not a collection of isolated points). As long as $\varphi > 0$ on $W_0$ the learning coefficient $\lambda$ is equal to a birational invariant of $W_0$ known in algebraic geometry as the \textbf{Real Log Canonical Threshold (RLCT)}. We will always assume this is the case, and now recall how $\lambda$ is described geometrically.

With $W_\epsilon = \{ w \in W : K(w) - K_0 \le \epsilon \}$ for sufficiently small $\epsilon$, \textbf{resolution of singularities }\citep{hironakaResolutionSingularitiesAlgebraic1964} gives us the existence of a birational proper map $g: M \to W_\epsilon$ from an analytic manifold $M$ which monomializes $K(w) - K_0$ in the following sense, described precisely in \citep[Theorem 6.5]{watanabeAlgebraicGeometryStatistical2009}: there are local coordinate charts $M_\alpha$ covering $g^{-1}(W_0)$ with coordinates $u$ such that the reparameterisation $w = g(u)$ puts $K(w) - K_0$ and $\varphi(w)dw$ into \emph{normal crossing form} 
\begin{gather}
    K(g(u)) - K_0 = u_1^{2k_1} \dots u_d^{2k_d} \label{eq:stdform1} \\
    |g'(u)| = b(u) u_1^{h_1} \dots u_d^{h_d} \label{eq:stdformg}\\
    \varphi(w) dw = \varphi(g(u)) | g'(u)|  du \label{eq:stdform2}
\end{gather}
for some positive smooth function $b(u)$. The RLCT of $K - K_0$ is independent of the (non-unique) resolution map $g$, and may be computed as \citep[Definition 6.4]{watanabeAlgebraicGeometryStatistical2009}
\begin{equation}\label{defn:lambda}
\lambda = \min_\alpha \min_{j = 1 \dots d} \frac{h_j + 1}{2k_j}.
\end{equation}
The multiplicity is defined as 
\begin{equation*}
m = \max_\alpha \# \{j: \lambda_j = \lambda\}.
\end{equation*}
For $P \in W_0$ there exist coordinate charts $M_{\alpha^*}$ such that $g(0) = P$ and \eqref{eq:stdform1}, \eqref{eq:stdformg} hold. The RLCT of $K - K_0$ at $P$ is then \citep[Definition 2.7]{watanabeAlgebraicGeometryStatistical2009}
\begin{equation}\label{defn:lambda_local}
\lambda(P) = \min_{\alpha^*} \min_{j = 1 \dots d} \frac{h_j + 1}{2k_j}\,.
\end{equation}
We then have the RLCT $\lambda = \inf_{P \in W_0} \lambda(P)$. In regular models, all $\lambda(P)$ are $d/2$ and hence the RLCT $\lambda=d/2$ and $m=1$, see \citep[Remark~1.15]{watanabeAlgebraicGeometryStatistical2009}. 

\subsection{Background assumptions in SLT}\label{appendix: slt assumptions}

There are a few technical assumptions throughout SLT that we shall collect in this section. We should note that these are only sufficient but not necessary conditions for many of the results in SLT. Most of the assumptions can be relaxed on a case by case basis without invalidating the main conclusions of SLT. For in depth discussion see \cite{watanabeAlgebraicGeometryStatistical2009, Watanabe2010-ud, watanabe2018}. 

With the same hypotheses as above, the log likelihood ratio 
\begin{equation*}
     r(x, w) := \log \frac{p(x |w_0)}{p(x |w)}
\end{equation*}
is assumed to be an $L^s(q(x))$-valued analytic function of $w$ with $s \geq 2$ that can be extended to a complex analytic function on $W_\mathbb{C} \subset \mathbb{C}^d$. A more conceptually significant assumption is the condition \textit{relatively finite variance}, which consists of the following two requirements:
\begin{enumerate}
    \item For any optimal parameters $w_1, w_2 \in W_0$, we have $p(x |w_1) = p(x |w_2)$ almost everywhere. This is also known as \textit{essential uniqueness}.
    \item There exists $c>0$ such that for all $w\in W$, we have
    \begin{align*}
        \mathbb{E}_{q(x)}\left[r(x, w)\right] \geq c \mathbb{E}_{q(x)}\left[r(x, w)^2\right]. 
    \end{align*}
\end{enumerate}

Note that if the true density $q(x)$ is \textit{realizable} by the model, i.e. there exist $w^* \in W$ such that $q(x) = p(x |w^*)$ almost everywhere, then these conditions are automatically satisfied. This includes settings for which the training labels are synthetically generated by passing inputs through a target model (in which case realizability is satisfied by construction), such as our experiments involving DLNs and some of the experiments involving MLPs. For experiments involving real data, it is unclear how reasonable it is to assume that relative finite variance holds.

\section{Well-definedness of the theoretical LLC}\label{appendix:well_definedness}

Here we remark on how we adapt the (global) learning coefficient of Definition \ref{defn:zeta_function} to define the local learning coefficient in terms of the poles of a zeta function, and how this relates to the asymptotic volume formula in Definition \ref{defn:llc} of the main text.

In addition to the setup described above, now suppose we have a local minimum $w^*$ of the negative log likelihood $L(w)$ and as in Section \ref{section:complexity_counting} we assume $B(w^*)$ to be a closed ball centered on $w^*$ such that $L(w^*)$ is the minimum value of $L$ on the ball. We define
\begin{gather*}
    V := \operatorname{Vol}( B(w^*) ) = \int_{B(w^*)} \varphi(w) dw\,,\\
    \bar \varphi(w) = \tfrac{1}{V} \varphi(w)
\end{gather*}
then we can form the \textbf{local triplet} $(p, q, \bar \varphi)$ with parameter space $B(w^*)$. Note that $W$ is cut out by a finite number of inequalities between analytic functions, and hence so is $B(w^*)$.

Provided $\varphi(w^*) > 0$ the prior does not contribute to the leading terms of the asymptotic expansions considered, and in particular does not appear in the asymptotic formula for the volume in Definition \ref{defn:llc}, and so we disregard it in the main text for simplicity.

Assuming relative finite variance ofr the local triplet, we can apply the discussion of Section \ref{appendix:slt_background}. In particular, we can define the LLC $\lambdawstar$ and the local multiplicity $m(\wstar)$ in terms of the poles of the ``local'' zeta function
\[
    \zeta(z, w^*) = \int_{B(w^*)} \left(K(w) - K(w^*)\right)^z \bar\varphi(w) dw
\]
Note that $L(w) - L(w^*) = K(w) - K(w^*)$ since $K,L$ differ by $S$ which does not depend on $w$. 

In order for the LLC to be a learning coefficient in its own right, it must be related, asymptotically, to the local free energy in the manner stipulated in \eqref{eq:localFn_nbhd_energy_entropy} in the main text. We verify this next. To derive the asymptotic expansion of the local free energy $F_n(B(w^*))$, we assume in addition that $\lambda(w^*) \le \lambda(P)$ if $P \in B(w^*)$ and $L(P) = L(w^*)$. That is, $w^*$ is at least as degenerate as any nearby minimiser. Note that the KL divergence for this triplet is just the restriction of $K: W \rightarrow \mathbb{R}$ to $B(w^*)$, but the local triplet has its own set of optimal parameters
\begin{equation}
W_{0}(w^*) = \{ w \in B(w^*) : L(w) = L(w^*) \}\,.
\end{equation}
Borrowing the proof in \citet[\S 3]{watanabeAlgebraicGeometryStatistical2009}, we can show that 
\begin{equation}
F_n(B(w^*)) = n L_n(w^*) + \lambda(w^*) \log n - (m(w^*) - 1) \log\log n + O_P(1) 
\label{eq:localFn_nbhd_asymptotics}
\end{equation}
where the difference between $\varphi$ and  $\bar \varphi$ contributes a summand $\log V$ to the constant term.  Note that the condition of relative finite variance is used to establish \eqref{eq:localFn_nbhd_asymptotics}. 
This explains why we can consider $\lambda(w^*)$ as a local learning coefficient, following the ideas sketched in \citep[Section 7.6]{watanabeAlgebraicGeometryStatistical2009}. The presentation of the LLC in terms of volume scaling given in the main text now follows from \citep[Theorem 7.1]{watanabeAlgebraicGeometryStatistical2009}.

\section{Reparameterization invariance of the LLC}
\label{appendix:reparam_invariance}


The LLC is invariant to \textit{local diffeomorphism} of the parameter space - roughly a locally smooth and invertible change of variables.

Note that this automatically implies several weaker notions of invariance, such as rescaling invariance in feedforward neural networks; other e.g. Hessian-based complexity measures have been undermined by their failure to stay invariant to this symmetry \citep{dinh2017}.
We now show how the LLC is invariant to local diffeomorphism. The fact that the LLC is an asymptotic quantity is crucial; this property would not hold for the volume $V(\epsilon)$ itself, for any value of $\epsilon$.

Let $U \subset W$ and $\widetilde{U} \subset \widetilde{W}$ be open subsets of parameter spaces $W$ and $\widetilde{W}$. A \textit{local diffeomorphism} is an invertible map $\phi: U \rightarrow \widetilde{U}$, such that both $\phi$ and $\phi^{-1}$ are infinitely differentiable. We further require that $\phi$ respect the loss function: that is, if $L: U \rightarrow \R$ and $\widetilde{L}: \widetilde{U} \rightarrow \R$ are loss functions on each space, we insist that $L(u) = \widetilde{L}(\phi(u))$ for all $u \in U$.

Supposing such a $\phi$ exists, the statement to be proved is that the LLC at $u^* \in U$ under $L(u)$ is equal to the LLC at $\widetilde{u}^* = \phi(u^*) \in \widetilde{U}$ under $\widetilde{L}(\widetilde{u})$. Define
\begin{align*}
V(\epsilon) = \int_{L(u) - L(u^*)<\epsilon} du, \quad \widetilde{V}(\epsilon) = \int_{\widetilde{L}(\widetilde{u}) - \widetilde{L}(\widetilde{u}^*)<\epsilon} d\widetilde{u}
\end{align*}

Now note that by the change of variables formula for diffeomorphisms, we have
\begin{equation*}
\widetilde{V}(\epsilon) = \int_{{L}(u) - L(u^*)<\epsilon} |\det D\phi(u)| du
\end{equation*}

where $\det D\phi(u)$ is the Jacobian determinant of $\phi$ at $u$.

The fact that $\phi$ is a local diffeomorphism implies that there exists constants $c_1, c_2$ such that $c_1 \leq |\det D\phi(u)| \leq c_2$ for all $u \in U$. This means that
\begin{equation*}
c_1 V(\epsilon) \leq \widetilde{V}(\epsilon) \leq c_2 V(\epsilon)
\end{equation*}

Finally, applying the definition of the LLC $\lambda$ and its multiplicity $m$, and leveraging the fact that this definition is asymptotic as $\epsilon \rightarrow 0$, we can conclude that
\begin{equation*}
V(\epsilon) \propto \widetilde{V}(\epsilon) \propto \epsilon^\lambda (-\log(\epsilon))^{m-1}
\end{equation*}

which demonstrates that the LLC is preserved by the local diffeomorphism $\phi$.

\section{Consistency of local WBIC}\label{appendix:consistency_local_WBIC}
In the main text, we introduced \eqref{eq:localWBIC} as an estimator of $\proxylocalFn$. Our motivation for this estimator comes directly from the well-known widely applicable Bayesian information criterion (WBIC) \citep{watanabeWidelyApplicableBayesian2013}. In this section, we refer to \eqref{eq:localWBIC} as the local WBIC and denote it $\wbic(\wstar)$

It is a direct extension of the proofs in \citet{watanabeWidelyApplicableBayesian2013} to show that the first two terms in the asymptotic expansion of the local WBIC match those of $F_n(\wstar,\gamma)$. By this we mean that it can be shown that
\begin{equation*}
F_n(\wstar,\gamma) = nL_n(\wstar) + \tilde \lambda(\wstar) \log n - (m-1) \log \log n + R_n
\end{equation*}
and
\begin{equation}
    \wbic(\wstar)= nL_n(\wstar) + \tilde \lambda(\wstar) \log n + U_n\sqrt{\tilde \lambda(\wstar) \log n/2} + O_P(1).
    \label{eq:local_wbic_expansion}
\end{equation}
This firmly establishes that \eqref{eq:localWBIC} is a good estimator of $\proxylocalFn$. However, it is important to understand that we cannot immediately conclude that it is a good estimator of $\nbhdLocalFn$.

We conjecture that $\tilde \lambda(\wstar)$ is equal to the LLC $\lambda(\wstar)$ given certain conditions on $\gamma$. So far, all our empirical findings suggest this. Detailed proof of this conjecture, in particular ascertaining the exact conditions on $\gamma$, will be left as future theoretical work. 

\section{Related work}
\label{appendix: related work}
We review both the singular learning theory literature directly involving the learning coefficient, as well as research from other areas of machine learning that may be relevant. This is an expanded version of the discussion in Section \ref{sec:related}.

\textbf{Singular learning theory}.
Our work builds upon the singular learning theory (SLT) of Bayesian statistics: good references are \citet{watanabeAlgebraicGeometryStatistical2009, watanabe2018}. The global learning coefficient, first introduced by \citep{watanabe2001}, provides the asymptotic expansion of the free energy, which is equivalent to the negative log Bayes marginal likelihood, an all-important important quantity in Bayesian analysis. Later work used algebro-geometric tools to bound or exactly calculate the learning coefficient for a wide range of machine learning models, including Boltzmann machines \cite{yamazaki2005a}, single-hidden-layer neural networks \cite{aoyagi2005}, DLNs \cite{aoyagi2024}, Gaussian mixture models \cite{yamazaki2003}, and hidden Markov models \cite{yamazaki2005b}.

SLT has also enhanced the understanding of model selection criteria in Bayesian statistics. Of particular relevance to this work is \citet{watanabeWidelyApplicableBayesian2013}, which introduced the WBIC estimator of free energy. This estimator has been applied in various practical settings (e.g. \citealp{endo2020, fontanesi2019, hooten2015, sharma2017, kafashan2021, semenova2020}). Some of the estimation methodology in this paper can be seen as a localized extension of the WBIC. Several other papers have explored improvements or alternatives to this estimator \cite{iriguchi2007, imai2019a, imai2019b}.

\textbf{Basin broadness}. The learning coefficient can be seen as a Bayesian version of \textit{basin broadness} measures, which typically attempt to empirically connect notions of geometric ``broadness" or ``flatness" with model complexity \cite{hochreiter1997, jiangFantasticGeneralizationMeasures2019}. However, the evidence supporting the (global) learning coefficient (in the Bayesian setting) is significantly stronger: the learning coefficient provably determines the Bayesian free energy to leading order \citet{watanabeAlgebraicGeometryStatistical2009}. We expect the utility of the learning coefficient as a geometric measure to apply beyond the Bayesian setting, but whether the connection with generalization will continue to hold is unknown.

\textbf{Neural network identifiability}. A core observation leading to singular learning theory is that the map $w \mapsto p(x | w)$ from parameters $w$ to statistical models $p(x | w)$ may not be one-to-one (in which case, the model is singular). This observation has been made in parallel by researchers studying \textit{neural network identifiability} \cite{sussmann1992, fefferman1994, krkov1994, phuong2020}. Recent work\footnote{Within the context of singular learning theory, this fact was known at least as early as \citet{fukumizu1996}.} has shown that the degree to which a network is identifiable (or inverse stable) is not uniform across parameter space \cite{berner2019, petersen2020, farrugiaroberts2023}. From this perspective, the LLC can be viewed as a quantitative measure of ``how identifiable" the network is near a particular parameter.

\textbf{Statistical mechanics of the loss landscape}. A handful of papers have explored related ideas from a statistical mechanics perspective. \citet{jules2023} use Langevin dynamics to probe the geometry of the loss landscape. \citet{zhangEnergyEntropyCompetition2018} show how a bias towards ``wide minima" may be explained by free energy minimization. These observations may be formalized using singular learning theory \cite{lamontCorrespondenceThermodynamicsInference2019}. In particular, the learning coefficient may be viewed as a heat capacity \cite{lamontCorrespondenceThermodynamicsInference2019}, and learning coefficient estimation corresponds to measuring the heat capacity by molecular dynamics sampling.

\textbf{Other model complexity measures.} The LLC is  set apart from a number of classic model complexity measures such as Rademacher complexity \citep{koltchinskii2000}, the VC dimension \citep{vapnik1971} because the latter measures act on an \textit{entire class of functions} while the LLC measures the complexity of a specific individual function within the context of the function class carved out by the model (e.g. via DNN architecture). This affords the LLC a better position for unraveling the theoretical mysteries of deep learning, which cannot be disentangled from the way in which DNNs are trained or the data that they are trained on. 

In the context studied here, our proposed LLC measures the complexity of a trained neural network rather than complexity over the entire function class of neural networks. It is also sensitive to the data distribution, making it ideal for understanding the intricate dance between function class, data properties, and implicit biases baked into different training heuristics.

Like earlier investigations by \citet{leBayesianPerspectiveGeneralization2018, zhangEnergyEntropyCompetition2018,lamontCorrespondenceThermodynamicsInference2019}, our notion of model complexity appeals to the correspondence between parameter inference and free energy minimization. This mostly refers to the fact that the posterior distribution over $w$ has high concentration around $\wstar$ if the associated local free energy, $F_n(B_\gamma(\wstar))$, is low. 
Viewed from the free energy lens, it is thus not surprising that ``flatter'' minima (low $\lambdawstar$) might be preferred over ``sharper'' minima (high $\lambdawstar$) even if the former has high training loss (higher $L_n(\wstar)$). Put another way, \eqref{eq:localFn_nbhd_energy_entropy} reveals that parameter inference is not about seeking the solution with the lowest loss. In the terminology of \citet{zhangEnergyEntropyCompetition2018}, parameter inference plays out as a competition between energy (loss) and entropy (Occam's factor). Despite spiritual similarities, our work starts to depart from that of \citet{zhangEnergyEntropyCompetition2018} and \citet{leBayesianPerspectiveGeneralization2018} in our technical treatment of the local free energy. These prior works rely on the Laplace approximation to arrive at the result
\begin{align}
    \nbhdLocalFn&= n L_n(\wstar) + \frac{d}{2} \log n + \frac{1}{2} \log \mathrm{det} H(w^*) + O_P(1),
    \label{eq:localFE_benign}
\end{align}
where $H$ is the Hessian of the loss. The flatness of a local minimum is calculated as $\log  \mathrm{det}  H(\wstar)$, which is, notably, ill-defined for neural networks\footnote{\citet{balasubramanianStatisticalInferenceOccam1997} includes all $O(1)$ terms in the Laplace expansion as a type of complexity measure.}. Indeed a concluding remark in \citet{zhangEnergyEntropyCompetition2018} points out that ``a more nuanced metric is needed to characterise flat minima with singular Hessian matrices." 
\citet{leBayesianPerspectiveGeneralization2018}, likewise, state in their introduction that ``to compute the (model) evidence, we must carefully account for this degeneracy", but then argues that degeneracy is not a major limitation to applying \eqref{eq:localFE_benign}. This is only partially true. For a very benign type of degeneracy, \eqref{eq:localFE_benign} is indeed valid. However, under general conditions, the correct asymptotic expansion of the local free energy is provided in \eqref{eq:localFn_nbhd_energy_entropy}. It might be said that while \citet{zhangEnergyEntropyCompetition2018} and \citet{leBayesianPerspectiveGeneralization2018} make an effort to account for degeneracies in the DNN loss landscape, they only take a small step up the degeneracy ladder while we take a full leap.

\textbf{Similarity to PAC-Bayes.} We have just described how the theoretical LLC is the sought-after notion of model complexity coming from earlier works who adopt the energy-entropy competition perspective. Interestingly, the actual LLC estimator also has connections to another familiar notion of model complexity. Among the diverse cast of complexity measures, see e.g., \citet{jiangFantasticGeneralizationMeasures2019} for a comprehensive overview of over forty complexity measures in modern deep learning, the LLC \textit{estimator} bears the most resemblance to PAC-Bayes inspired flatness/sharpness measures \citep{neyshabur2017}. Indeed, it may be immediately obvious that, other than the scaling of $n \betastar$, $\hatlambdawstar$ can be viewed as a PAC-Bayes flatness measure which utilises a very specific posterior distribution localised to $\wstar$. Recall the canonical PAC-Bayes flatness measure is based on 
\begin{equation}
\lambda_{\mathrm{PAC-bayes}}(\wstar) = \E_{q(w|\wstar)} \ell_n(w) - \ell_n(\wstar),    
\label{eq:pacbayes_flatness}
\end{equation}
where $\ell_n$ is a general empirical loss function (which in our case is the sample negative log likelihood) and the ``posterior"
distribution $q$ is often taken to be Gaussian, i.e., $q(w|\wstar) = \mathcal N(\wstar, \sigma^2 I)$. A simple derivation shows us that the quantity in \eqref{eq:pacbayes_flatness}, if we use a Gaussian $q$ around $\wstar$, reduces approximately to
$
\frac{1}{2} \sigma^2 Tr(H(\wstar)),
$
where $H$ is the Hessian of the loss, i.e., $H(\wstar) = \nabla^2_w \ell_n(w) |_{\wstar}$. 
However, for singular models, the posterior distribution around $\wstar$, e.g., \eqref{eq:local_tempered_posterior}, is decidedly \textit{not} Gaussian. This calls into question the standard choice of the Gaussian posterior in \eqref{eq:pacbayes_flatness}. 

\textbf{Learning capacity.} Finally, we briefly discuss concurrent work that measures a quantity related to the learning coefficient. In \citet{chenLearningCapacityMeasure2023}, a measure called the learning capacity is proposed to estimate the complexity of a hypothesis class. The learning capacity can be viewed as a finite-$n$ version of the learning coefficient; the latter only appears in the $n \to \infty$ limit. \citet{chenLearningCapacityMeasure2023} is largely interested in the learning capacity as a function of training size $n$. They discover the learning capacity saturates at very small and large $n$ with a sharp transition in between. 

\textbf{Applications.} Recently, the LLC estimation method we introduce here has been used to empirically detect ``phase transitions" in toy ReLU networks \cite{chenLearningCapacityMeasure2023}, and the development of in-context learning in transformers \cite{hoogland2024}.

\section{Model complexity vs model-independent complexity} \label{appendix: complexity disambiguation}

In this paper, we have described the LLC as a measure of ``model complexity." It is worth clarifying what we mean here --- or rather, what we \textit{do not} mean. This clarification is in part a response to \citet{skalse2023}.

We distinguish measures of ``model complexity," such as those traditionally found in statistical learning theory, from measures of ``model-independent complexity," such as those found in algorithmic information theory. Measures of model complexity, like the parameter count, describe the expressivity or degrees of freedom available to a \textit{particular model}. Measures of model-independent complexity, like Kolmogorov complexity, describe the complexity inherent to the \textit{task itself}.

In particular, we emphasize that --- a priori --- the LLC is a measure of model complexity, \textit{not} model-independent complexity. It can be seen as the amount of information required to nudge a model towards $w^*$ and away from other parameters. Parameters with higher LLC are more complex for \textit{that particular model} to implement. 

Alternatively, the model is \textit{inductively biased}\footnote{The role of the LLC in inductive biases is only rigorously established for Bayesian learning, but we suspect it also applies for learning with SGD.} towards parameters with lower LLC --- but a different model could have different inductive biases, and thus different LLC for the same task. This is why it is not sensible to conclude that a bias towards low LLC, would, on its own, explain observed ``simplicity bias" in neural networks \cite{valle2018} --- this is tautological, as \citet{skalse2023} noted.

To highlight this distinction, we construct a statistical model where these two notions of complexity diverge. Let $f_1(x)$ be a Kolmogorov-simple function, like the identity function. Let $f_2(x)$ be a Kolmogorov-complex function, like a random lookup table. Then consider the following regression model with a single parameter $w \in [0, 1]$:
\begin{equation*}
f(x, w) = w^8 f_1(x) + (1 - w^8) f_2(x)
\end{equation*}
For this model, $f_1(x)$ has a learning coefficient of $\lambda = \frac{1}{2}$, whereas $f_2(x)$ has a learning coefficient of $\lambda = \frac{1}{16}$. Therefore, despite $f_1(x)$ being more Kolmogorov-simple, it is more complex for $f(x, w)$ to implement --- the model is biased towards $f_2(x)$ instead of $f_1(x)$, and so $f_1(x)$ requires relatively more information to learn.

Yet, this example feels contrived: in realistic deep learning settings, the parameters $w$ do not merely interpolate between handpicked possible algorithms, but themselves \textit{define} an internal algorithm based on their values. That is, it seems intuitively like the parameters play a role closer to ``source code" than ``tuning constants."

Thus, while in general LLC is not a model-independent complexity measure, it seems distinctly possible that for neural networks (perhaps even models in some broader ``universality class"), the LLC could be model-independent in some way. This would theoretically establish the inductive biases of neural networks. We believe this to be an intriguing direction for future work.

\section{SGLD-based LLC estimator: minibatch version pseudocode}\label{appendix:minibatch_hatlambdawstar}

\begin{algorithm}[h]
    \caption{computing $\hatlambdawstar$}
    \label{alg:hatlambda_code}
    \textbf{Input}: 
    \begin{itemize}
        \item initialization point: $w^*$
        \item scale: $\gamma$
        \item step size: $\epsilon$
        \item number of iterations: SGLD\_iters 
        \item batch size: $m$
        \item dataset of size $n$: $\mathcal{D}_n = \{(x_i, y_i)\}_{i = 1, \dots, n}$
        \item averaged log-likelihood function for $w \in \mathbb R^d$ and arbitrary subset $D$ of data: 
        \begin{equation*}
        \mathrm{logL}(D, w) = \frac{1}{\abs{D}}\sum_{(x_i, y_i) \in D} \log p(y_i|x_i, w)
        \end{equation*}
    \end{itemize}
    \textbf{Output}: $\hatlambdawstar$
    \begin{algorithmic}[1]
        \STATE $\beta^* \gets \frac{1}{\log n}$ \COMMENT{Optimal sampling temperature.}
        \STATE $w \gets w^*$ \COMMENT{Initialize at the given parameter}
        \STATE $\mathrm{arrayLogL} \gets [\,\,]$
        \FOR{$t = 1 \dots \text{SGLD\_iters}$}
            \STATE $B \gets$ random minibatch of size $m$
            \STATE append $\mathrm{logL}(B, w)$ to $\mathrm{arrayLogL}$
            \STATE $\eta \sim N(0, \epsilon)$ \COMMENT{$d$-dimensional Gaussian, variance $\epsilon$}
            \STATE $\Delta w \gets \frac{\epsilon}{2} \left [ \gamma (w^* - w) + n\betastar \nabla_w \mathrm{logL}(B, w) \right] + \eta$
            \STATE $w \gets w + \Delta w$
        \ENDFOR
        \STATE $\widehat{\wbic} \gets -n \cdot \mathrm{Mean}(\mathrm{arrayLogL})$
        \STATE $nL_n(w^*) \gets -n \cdot \mathrm{logL}(\mathcal{D}_n, w^*)$
        \STATE $\hatlambdawstar \gets \frac{\widehat{\wbic} - nL_n(w^*)}{\log n}$
        \STATE \textbf{return} $\hatlambdawstar$
    \end{algorithmic}
\end{algorithm}

\clearpage
\section{Recommendations for accurate LLC estimation and troubleshooting}\label{appendix:recommendations}

In this section, we collect several recommendations for estimating the LLC accurately in practice. Note that these recommendations are largely based on our experience with LLC estimation for DLN (see Section \ref{section:expt_dln}) as it is the only realistic model (it being wide and deep) where the theoretical learning coefficient is available.

\subsection{Step size}

From experience, the most important hyperparameter to the performance and accuracy of the method is the step size $\epsilon$. If the step size is too low, the sampler may not equilibriate, leading to underestimation. If the step size is too high, the sampler can become numerically unstable, causing overestimation or even ``blowing up" to \texttt{NaN} values.

Manual tuning of the step size is possible. However we strongly recommend a particular diagnostic based on the acceptance criterion for Metropolis-adjusted Langevin dynamics (MALA). This is used to \textit{correct} numerical errors in traditional MCMC, but here we use it only to \textit{detect} them.

In traditional (full-gradient) MCMC, numerical errors caused by the step size are completely corrected by a secondary step in the algorithm, the \textit{acceptance check} or \textit{Metropolis correction}, which accepts or rejects steps with some probability roughly\footnote{Technically, the acceptance probability is based on maintaining detailed balance, not necessarily numerical error, as can be seen in the case of e.g. Metropolis-Hastings. But this is a fine intuition for gradient-based algorithms like MALA or HMC.} based on the likelihood of numerical error. The proportion of steps accepted additionally becomes an important diagnostic as to the health of the algorithm: a low acceptance ratio indicates that the acceptance check is having to compensate for high levels of numerical error.

The acceptance probability between step $X_k$ and proposed step $X_{k+1}$ is calculated as:
\begin{equation*}
\min\left(1, \frac{\pi(X_k)\,q(X_k |X_{k+1})}{\pi(X_{k+1})\,q(X_{k+1} |X_k)}\right)
\end{equation*}
where $\pi(x)$ is the probability density at $x$ (in our case, $\log \pi(x) = \beta n L_n(x)$), and $q(x' |x)$ is the probability of our sampler transitioning from $x$ to $x'$.

In the case of MALA, $q(x' |x) \not= q(x |x')$ and so we must explicitly calculate this term. For MALA, it is:
\begin{equation*}
q(x' |x) \propto \exp(-\frac{1}{4\epsilon} ||x' - x - \epsilon \nabla \log \pi(x)||^2)
\end{equation*}
We choose to use MALA's formula because we are using SGLD, and both MALA and SGLD propose steps using Langevin dynamics. MALA's formula is the correct one to use when attempting to apply Metropolis correction to Langevin dynamics.

For various reasons, directly implementing such an acceptance check for stochastic-gradient MCMC (while possible) is typically either ineffective or inefficient. Instead we use the acceptance probability merely as a diagnostic.

\textit{We recommend tuning the step size such that the average acceptance probability is in the range of 0.9-0.95}. Below this range, increase step size to avoid numerical error. Above this range, consider decreasing step size for computational efficiency (to save on the number of steps required). For efficiency, we recommend calculating the acceptance probability for only a fraction of steps --- say, one out of every twenty.

Note that since we are not actually using an acceptance check, these acceptance ``probabilities" are not really probabilities, but merely diagnostic values.

\subsection{Step count and burn-in}

The step count for sampling should be chosen such that the sampler has time to equilibriate or ``burn in''. Insufficient step count may lead to underestimating the LLC. Excessive step count will not degrade accuracy, but is unnecessarily time-consuming.

\begin{figure}[ht]
\begin{center}
\centerline{\includegraphics[width=\textwidth]{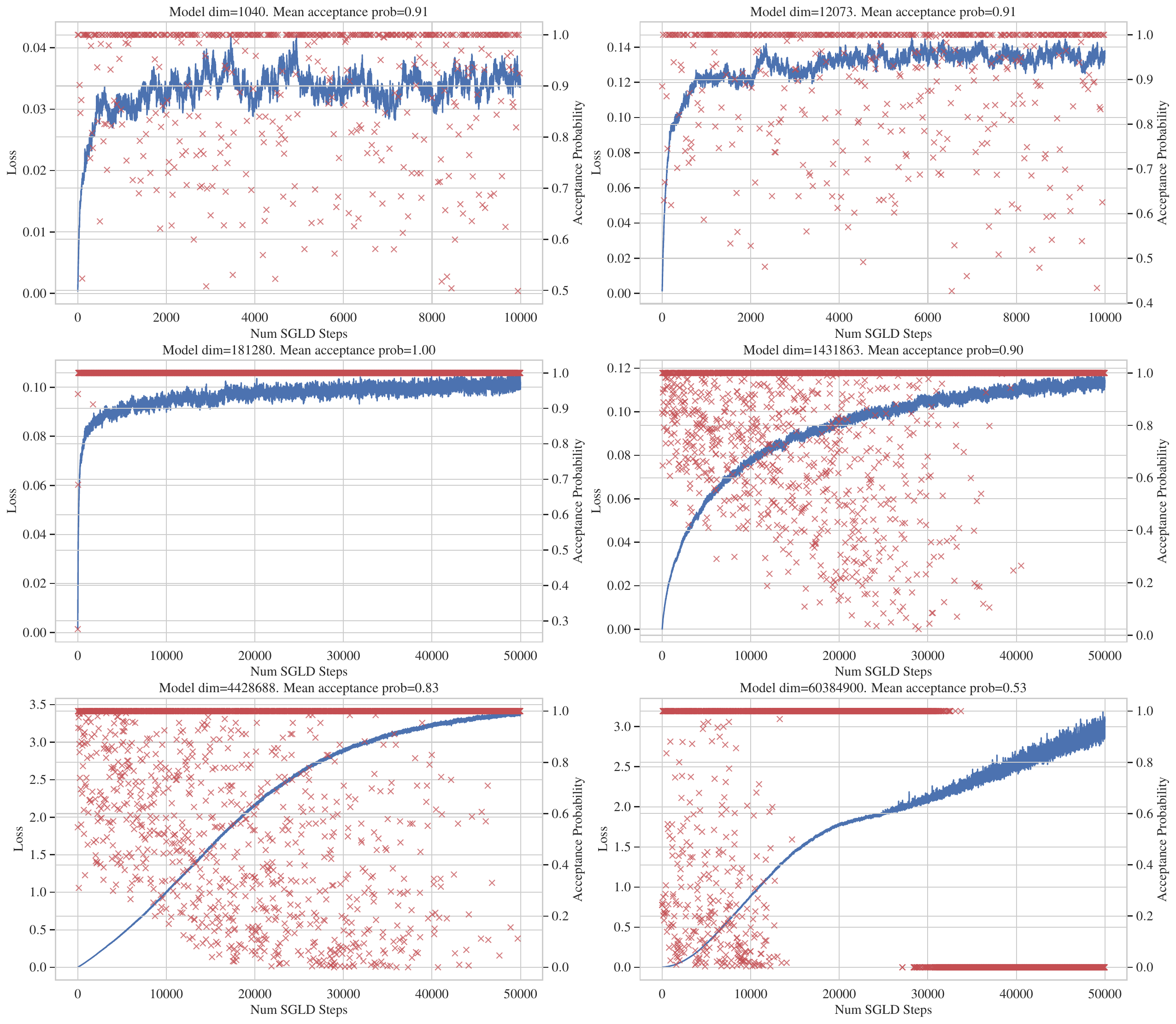}}
\caption{Sample loss trace (blue, left axis) and MALA acceptance probability (red, right axis) over DLN training trajectories at different model sizes.}
\label{fig:loss trace}
\end{center}
\end{figure}

\textit{We recommend increasing the number of steps until the loss, $L_m(w_t)$, stops increasing after some period of time.} This can be done with manual inspection of the loss trace. See Figure \ref{fig:loss trace} for some examples of loss trace and MALA acceptance probability over SGLD trajectories for DLN model at different scale. 

It is worth noting that the loss trace should truly be flat --- a slow upwards slope can still be indicative of significant underestimation.

\textit{We also recommend that samples during this burn-in period are discarded.} That is, loss values should only be tallied once they have flattened out. This avoids underestimation.

\subsection{Other issues and troubleshooting} \label{appendix: sgld troubleshooting}
We note some miscellaneous other issues and some troubleshooting recommendations:  
\begin{itemize}
    \item \textbf{Negative LLC estimates}: This can happen when $w^*$ fails to be near a local minimum of $L(w)$. However even when $w^*$ is a local minimum, we still might get negative LLC estimates if we are not careful. This can happen when the SGLD trajectory wanders to an area with lower loss than the initialization, causing the numerator in \eqref{eq:hatlambda} to be negative. This could be alleviated by smaller step size or shorter chain length. This, however, risks under-exploration. This can also be alleviated by having larger restoring force $\gamma$. This risks the issue discussed below. 
    \item \textbf{Large $\gamma$}: An overly concentrated localizing prior ($\gamma$ too large) can overwhelm the gradient signal coming from the log-likelihood. This can result in samples that are different from the posterior, destroying SGLD's sensitivity to the local geometry. 
\end{itemize}

To sum up, in pathological cases like having SGLD trajectory falling to lower loss region or blowing up beyond machine floating number limits, we recommend keeping $\gamma$ small (1.0 to 10.0), gradually lowering the step-size $\epsilon$ while lengthening the sampling chain so that the loss trace still equilibrates.

\clearpage
\section{Learning coefficient of DLN \citep{aoyagi2024}}\label{appendix: aoyagi dln lambda}
A DLN is a feedforward neural network without nonlinear activation. Specifically, a biasless DLN with $M$ hidden layers, layer sizes $H_1, H_2, \dots, H_M$ and input dimension $H_0$ is given by:
\begin{equation}\label{eq:dln def}
  y = f(x, w) = W_M \ldots W_2 W_1 x
\end{equation}
where $x \in \R^{H_0}$ is the input vector and the model parameter $w$ consist of the weight matrices $W_j$ of shape $H_{j} \times H_{j-1}$ for $j = 1, \dots, M$. 
Given a DLN, $f(x, w)$, (c.f. Equation \ref{eq:dln def}) with $M$ hidden layers, layer sizes $H_1, H_2, \dots, H_M$ and input dimension $H_0$, the associated regression model with additive Gaussian noise is given by 
\begin{align}
    p(x, y |w) = \frac{q(x)}{\sqrt{2\pi \sigma^2}^{H_M}} e^{-\frac{1}{2\sigma^2} \| y - W_M \dots W_2 W_1 x \|^2}
\end{align}
where $q(x)$ is some distribution on the input $x \in \R^{H_0}$, $w = (W_1, \dots, W_M)$ is the parameter consisting of the weight matrices $W_j$ of shape $H_{j} \times H_{j-1}$ for $j = 1, \dots, M$ and $\sigma^2$ is the variance of the additive Gaussian noise. Let $q(x, y)$ be the density of the true data generating process and $w^* = (W_1^*, \dots, W_M^*)$ be an optimal parameter that minimizes the KL-divergence between $q(x, y)$ and $p(x, y |w)$. 

Here we shall pause and emphasize that this result gives us the (global) learning coefficient, which is conceptually distinct from the LLC. They are related: the learning coefficient is the minimum of the LLCs of the \textit{global} minima of the population loss. In our experiments, we will be measuring the LLC at a randomly chosen global minimum of the population loss. While we don't expect LLCs to differ much among global minima for DLN, we do not know that for certain and it is of independent interest that the estimated LLC can tell us about the learning coefficient. 

\begin{theorem}[DLN learning coefficient, \citealp{aoyagi2024}]
    
    Let $r := \mathrm{rank}\brac{W_M^* \dots W_2^* W_1^*}$ be the rank of the linear transformation implemented by the true DLN, $f(x, w)$ and set $\Delta_j := H_j - r$, for  $j = 0, \dots, M$. There exist a subset $\Sigma \subset \{0, 1, \dots, M\}$ of indices, $\Sigma = \{\sigma_1, \dots, \sigma_{\ell + 1}\}$ with cardinality $\ell + 1$ that satisfy the following conditions:
    \begin{align*}
        \max\{\Delta_{\sigma} \mid \sigma \in \Sigma\} &< \min\{\Delta_{k} \mid k \not \in \Sigma \} \\
        \sum_{\sigma \in \Sigma} \Delta_{\sigma} &\geq \ell \cdot \max\{\Delta_{\sigma} \mid \sigma \in \Sigma\} \\
        \sum_{\sigma \in \Sigma} \Delta_{\sigma} &< \ell \cdot  \min\{\Delta_{\sigma} \mid \sigma \not \in \Sigma\}.
    \end{align*}
    Assuming that the DLN truth-model pair $\brac{q(x, y), p(x, y |w)}$ satisfies the relatively finite variance condition (Appendix \ref{appendix: slt assumptions}), their learning coefficient is then given by
    \begin{align*}
        \lambda = \frac{-r^2 + r (H_0 + H_L)}{2} + \frac{a(\ell - a)}{4\ell} - \frac{\ell (\ell -1)}{4} \brac{\frac{1}{\ell} \sum_{j = 1}^{\ell + 1}\Delta_{\sigma_j}}^2 + \frac{1}{2} \sum_{1 \leq i < j \leq \ell + 1} \Delta_{\sigma_i} \Delta_{\sigma_j}. 
    \end{align*}
\end{theorem}

As we mention in the introduction, trained neural networks are less complex than they seem. It is natural to expect that this, if true, is reflected in deep networks having more degenerate (good parameters has higher volume) loss landscape.  With the theorem above, we are given a window into the volume scaling behaviour in the case of DLNs, allowing us to investigate an aspect of this hypothesis.  

\begin{figure}[ht]
\begin{center}
\centerline{\includegraphics[width=0.8\textwidth]{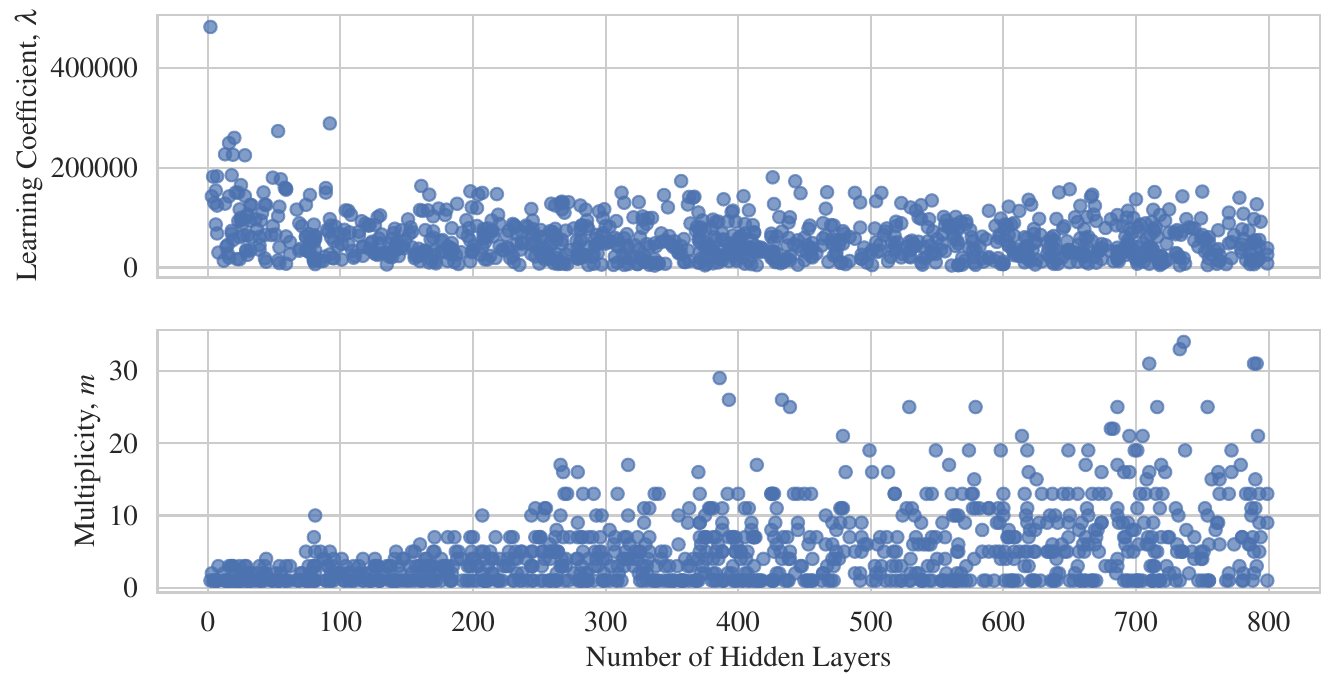}}
\caption{The top graph shows $\lambda$ decreasing as the DLN becomes deeper, even though model parameter count increases with number of layers. The bottom graph shows the true multiplicities, $m$. Since regular models can only have  $m = 1$, the graph shows that most of these randomly generated DLNs are singular. }
\label{fig:numlayer vs lambda}
\end{center}
\end{figure}

Figure \ref{fig:numlayer vs lambda} shows the the true learning coefficient, $\lambda$, and the multiplicity, $m$, of many randomly drawn DLNs with different numbers of hidden layers. Observe that $\lambda$ decreases with network depth.

This plot is generated by creating networks with 2-800 hidden layers, with width randomly drawn from 100-2000 including the input dimension. The overall rank of the DLN is randomly drawn from the range of zero to the maximum allowed rank, which is the minimum of the layer widths. See also Figure 1 in \cite{aoyagi2024} for more theoretical examples of this phenomenon. 

\clearpage
\section{Experiment: DLN} \label{appendix:expt_dln}

We compare the estimated $\hat{\lambda}(w^*)$ against theoretical $\lambda$ (with $w^*$ being a randomly generated true parameter), by randomly generating many DLNs with different architectures and model sizes that span several orders of magnitude (OOM). Each DLN is constructed randomly, as follows. Draw an integer $M \sim U(M_{low}, \dots,  M_{high})$ as the number of hidden layers, where $U(a, \dots, b)$ denotes the discrete uniform distribution on the finite set $\{a, a + 1, \dots, b\}$. Then, draw layer size $H_j \sim U(H_{low}, \dots, H_{high})$ for each $j = 0, \dots, M$ where $H_0$ denotes the input dimension. The weight matrix $W_j$ for layer $j$ is then a $H_{j} \times H_{j -1}$ matrix with each matrix element independently sampled from $N(0, 1)$ (random initialization). To obtain a more realistic true parameter, with probability $0.5$, each matrix $W_j^*$ is modified to have a random rank of $r \sim U(0, \dots, \min(H_{j-1}, H_j))$. For each DLN generated, a corresponding synthetic training dataset of size $n$ is generated to be used in SGLD sampling. 

The configuration values $M_{low}, M_{high}, H_{low}, H_{high}$ are chosen differently for separate set of experiments with model size targeting DLN size of different order of magnitude. See Table \ref{tab:dln experiment param} for the values used in the experiments. SGLD hyperparameters $\epsilon$, $\gamma$, and number of steps are chosen to suit each set of experiments according to our recommendations outlined in Appendix \ref{appendix:recommendations}. The exact hyperparameter values are given in the following section. 

\textit{We emphasize that hyperparameters must be tuned independently for different model sizes.} In particular, the required step size for numerical stability tends to \textit{decrease} for larger models, forcing a compensatory \textit{increase} in the step count. See Table \ref{tab:dln experiment param} for an example of this tuning with scale.

Future work using e.g. $\mu$-parameterization \cite{yang2022} may be able to alleviate this issue.

\subsection{Hyperparameters and further details}

As described in the prior section, the experiments shown in Figure \ref{fig:scalability} consist of randomly constructed DLNs. For each target order of magnitude of DLN parameter count, we randomly sample the number of layers and their widths from a different range. We also use a different set of SGLD hyperparameter chosen according to the recommendation made in Section \ref{appendix:recommendations}. Configuration values that varies across OOM are shown in Table \ref{tab:dln experiment param} and other configurations are as follow
\begin{itemize}
    \item Batch size used in SGLD is 500. 
    \item The amount of burn-in steps used for SGLD samples is set to 90\% of total SGLD chain length, i.e. only the last 10\% of SGLD samples are used in estimating the LLC. 
    \item The parameter $\gamma$ is set to $1.0$. 
    \item For each DLN, $f(x, w^*)$ with a chosen true parameter $w^*$, a synthetic dataset, $\{(x_i, y_i)\}_{i = 1, \dots, n}$ is generated by randomly sampling each element of the input vector $x$ uniformly from the interval $[-10, 10]$ and set the output as $y = f(x, w^*)$, which effectively means we are setting a very small noise variance $\sigma^2$. 
    \item For LLC estimation done at a trained parameter instead of the true parameter (shown in right plot in Figure \ref{fig:scalability}), the network is first trained using SGD with learning rate 0.01 and momentum 0.9 for 50000 steps. 
\end{itemize}
For each target OOM, a number of different experiments are run with different random seeds. The number of such experiment is determined by our compute resources and is reported in Table \ref{tab:dln experiment param} with some experiment failing due to SGLD chains ``blowing up'' (See discussion in Appendix \ref{appendix:recommendations}) for the SGLD hyperparameters used. Figure \ref{fig:mean mala probs histogram} shows that there is a left tail to the mean MALA acceptance rate distribution that hint at instability in SGLD chains encountered in some $\hat{\lambda}(w^*)$ estimation run.

\begin{table}
    \centering
    \begin{tabular}{c|p{2cm} | p{2cm} | ccccccc}
        \toprule
         \textbf{OOM}  & \textbf{Num layers}, $M_{low}$-$M_{high}$ & \textbf{Widths}, $H_{low}$-$H_{high}$   & $\epsilon$         & \textbf{Num SGLD steps} & $n$ & \textbf{Num experiments}\\
         \midrule
         1k   &  2-5       & 5-50      & $5 \times 10^{-7}$ &  10k  & $10^5$ & 99 \\
         10k  &  2-10      & 5-100     & $5 \times 10^{-7}$ &  10k  & $10^5$ & 100 \\
         100k &  2-10      & 50-500    & $1 \times 10^{-7}$ &  50k  & $10^6$ & 100 \\
         1M   &  5-20      & 100-1000  & $5 \times 10^{-8}$ &  50k  & $10^6$ & 99 \\
         10M  &  2-20      & 500-2000  & $2 \times 10^{-8}$ &  50k  & $10^6$ & 93 \\
         100M &  2-40      & 500-3000  & $2 \times 10^{-8}$ &  50k  & $10^6$ & 54 \\
         \bottomrule
    \end{tabular}
    \vskip .2em
    \caption{Table of experimental configuration for each batch of experiment at different order of magnitudes (OOM) in DLN model size. $n$ denotes the training dataset size and $\epsilon$ denotes SGLD step size.}
    \label{tab:dln experiment param}
\end{table}

\subsection{Additional plots for DLN experiments}
\begin{figure}[!ht]
\begin{center}
\centerline{\includegraphics[width=\textwidth]{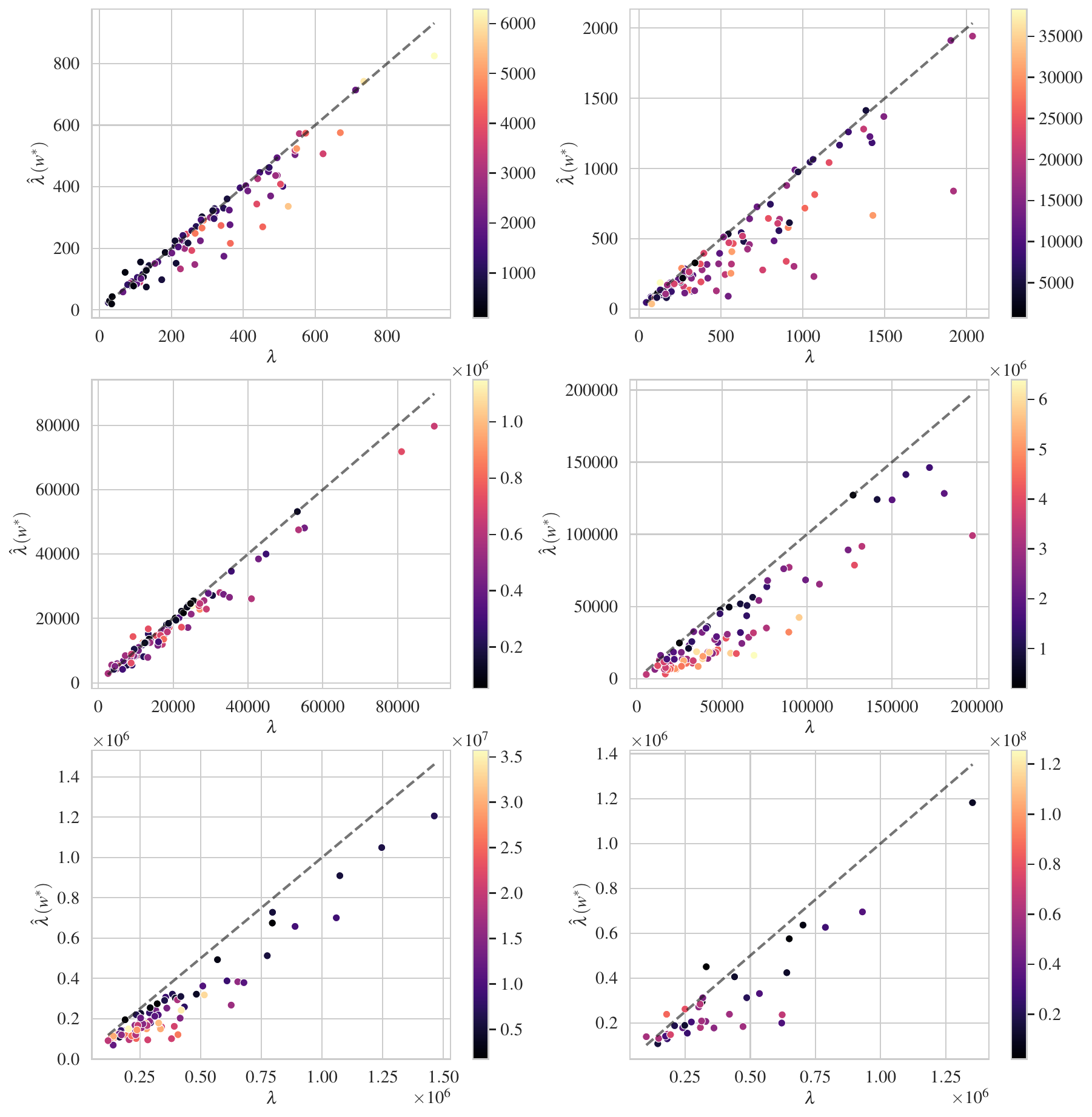}}
\caption{Supplementary plot to Figure \ref{fig:scalability}. Each plot shows a single batch of DLN experiment with model size at different order of magnitude. The SGLD hyperparameter is tuned once for each batch. Their values are listed in Table \ref{tab:dln experiment param} In contrast to Figure \ref{fig:scalability} which is in log scale, all plots here are in linear scale.}
\label{fig:lambdahat-vs-lambda linear scale}
\end{center}
\end{figure}

\begin{figure}[!ht]
\begin{center}
\centerline{\includegraphics[width=0.75\textwidth]{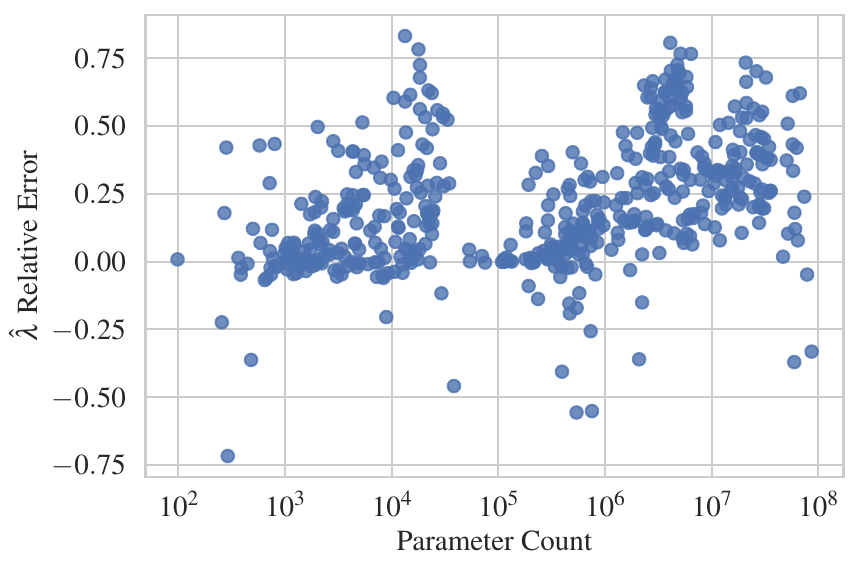}}
\caption{Relative error of estimated LLC compared to the theoretical learning coefficient, for DLNs across different orders of magnitude of model size.}
\label{fig:lambdahat relative error}
\end{center}
\end{figure}

\begin{figure}[!ht]
\begin{center}
\centerline{\includegraphics[width=0.75\textwidth]{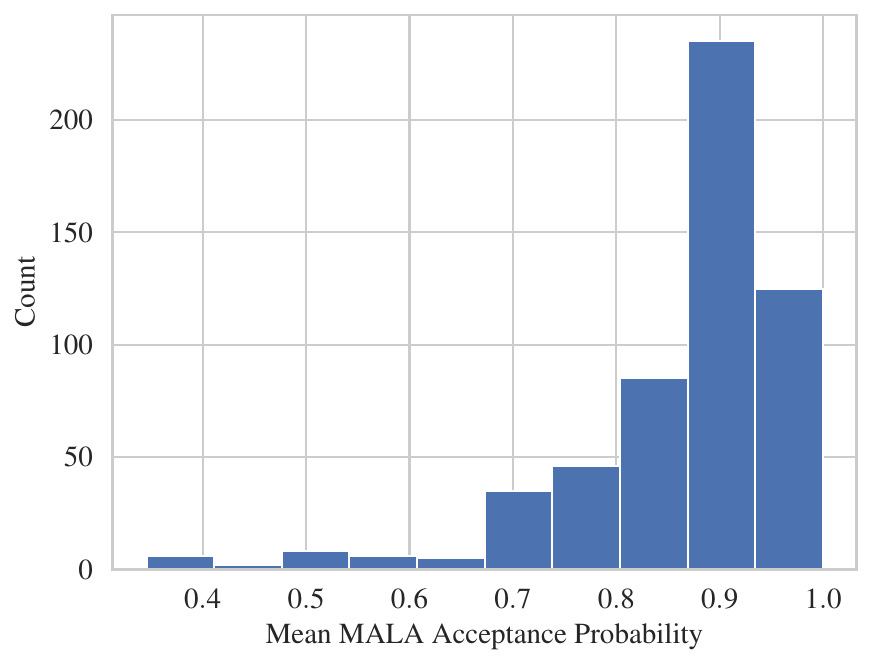}}
\caption{Mean MALA acceptance probability over the entire SGLD trajectory for every DLN experiment. Model size is not the only factor affecting the correct scale for SGLD step size. Local geometry varies significantly among different models, and among different neighbourhoods in the parameter space. Without tuning SGLD hyperparameters individually for each experiment, we get a spread of (mean) MALA acceptance probability over all experiments. Those with low acceptance probability may indicate poor $\hat{\lambda}(w^*)$ estimation quality. }
\label{fig:mean mala probs histogram}
\end{center}
\end{figure}

\begin{itemize}
    \item Figure \ref{fig:lambdahat-vs-lambda linear scale} is a linear scale version of Figure \ref{fig:scalability} in the main text. This shows the estimated LLC against the true learning coefficients for experiments at different model size range without log-scale distortion. 
    \item Figure \ref{fig:lambdahat relative error} shows the relative error $(\lambda - \hat{\lambda}(w^*)) / \lambda$ across multiple orders of magnitude of DLN model size. 
\end{itemize}

\clearpage
\section{Experiment: LLC for ResNet}
\label{appendix:expt_cifar}
This section provides details for the experiments where we aimed to investigate how LLC vary over neural network training with different training configuration. These experiments parallel those performed in \citet{Dherin2022-sa} who propose the\textit{geometric complexity} measure.

We train ResNet18 \citep{He2016-cn} on the CIFAR10 dataset \citep{Krizhevsky2009-wh} using SGD with cross-entropy loss. We vary SGD hyperparameters such as learning rate, batch size, momentum and $L^2$-regularization rate and track the resulting LLC estimates over evenly spaced checkpoints of SGD iterations. 

For the LLC estimates to be comparable, we need to ensure that the SGLD hyperparameters used in the LLC algorithm is the same. To this end, for every set of experiments where we vary a single optimizer hyperparameter, we first perform a calibration run to select SGLD hyperparameters according to the recommendation out lined in Appendix \ref{appendix:recommendations}. Once selected, this set of SGLD hyperparameter is then used for all LLC estimation within the set of experiments. That include the LLC estimation for every checkpoint of every ResNet18 training run for different optimizer configuration. 
Also following the same recommendation, we also burn away 90\% of the sample trajectory and only use last 10\% of samples. We also note that, since the SGLD hyperparameters are not tuned for all experiments within a single set, there are possibilities of negative LLC estimates or divergent SGLD chains. See Appendix \ref{appendix: sgld troubleshooting} for discussion on such cases and how to troubleshoot them. We manually remove these cases. They are rare enough that they do not change the LLC curves, just widen the error bars (confidence intervals) due to having less repeated experiments. 



Each experiment is repeated with 5 different random seeds. While the model architecture and dataset (including the train-test split) is fixed, other factors like the network initialization, training trajectories and SGLD samples are randomized. In each plot, the error bars show the 95\% confidence intervals over 5 repeated experiments of the statistics plotted. The error bars were calculated using the inbuilt error bar function to Python Seaborn library \citep{Waskom2021} plotting function, \texttt{seaborn.lineplot(..., errorbar=("ci", 95), ...)}.

\subsection{Details for main text Figures \ref{fig:implicit_regularization}}

For experiments that vary the learning rate in Figure \ref{fig:implicit_regularization} (top), for each learning rate value in $[0.005, 0.05, 0.01, 0.1, 0.2]$ we run SGD \textit{without} momentum with a fixed batch size of 512 for 30000 iterations. LLC estimations were performed every 1000 iterations with SGLD hyperparameters as follow: step size $\epsilon = 2 \times 10^{-7}$, chain length of 3000 iterations, batch size of 2048 and  $\gamma = 1.0$.

For experiments that vary the batch size in Figure \ref{fig:implicit_regularization} (middle), for each batch size value in $[16, 32, 64, 128, 256, 512, 1024]$ we run SGD \textit{without} momentum with a fixed learning rate of 0.01 for 100000 iterations. LLC estimations were performed every 2500 iterations with SGLD hyperparameters as follow: step size $\epsilon = 2 \times 10^{-7}$, chain length of 2500 iterations, batch size of 2048 and  $\gamma = 1.0$.

For experiments that vary the SGD momentum in Figure \ref{fig:implicit_regularization} (bottom), for each momentum value in $[0.0, 0.1, 0.2, 0.3, 0.4, 0.5, 0.6, 0.7, 0.8, 0.9]$ we run SGD with momentum with a fixed learning rate of 0.05 and a fixed batch size of 512 for 20000 iterations. LLC estimations were performed every 1000 iterations with SGLD hyperparameters as follow: step size $\epsilon = 2 \times 10^{-7}$, chain length of 3000 iterations, batch size of 2048 and  $\gamma = 1.0$.

\subsection{Additional ResNet18 + CIFAR10 LLC experiments}

\begin{figure}[ht]
    \centering
    \includegraphics[width=0.99\textwidth, keepaspectratio]{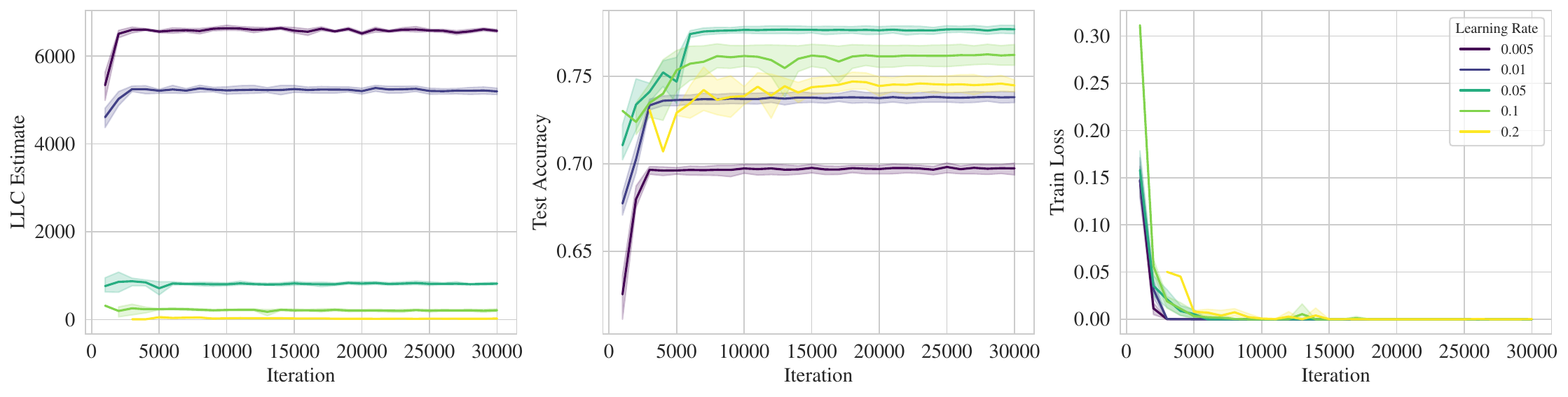} \\
    \includegraphics[width=0.99\textwidth, keepaspectratio]{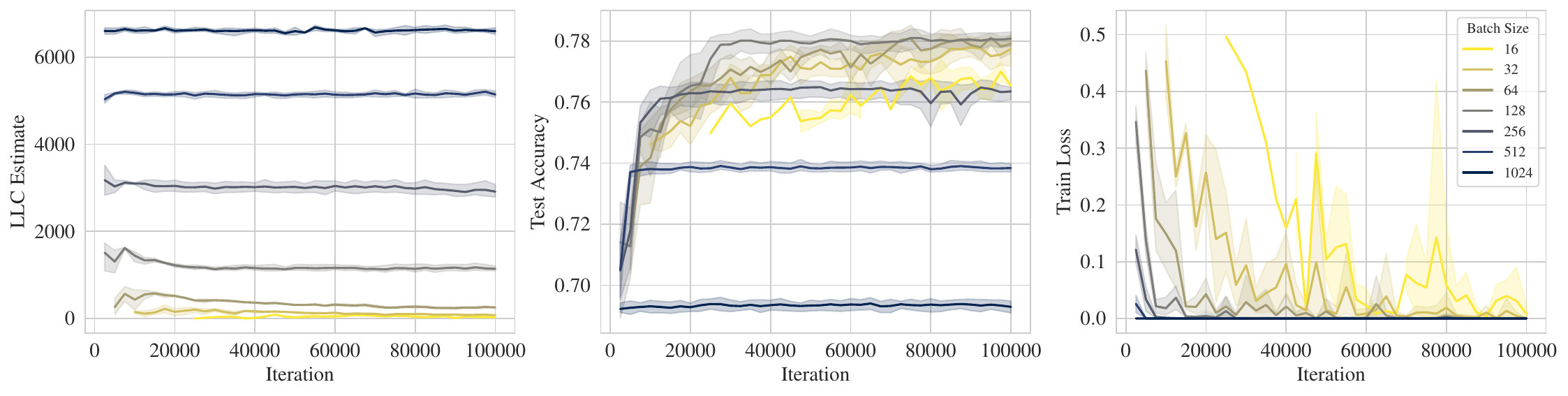} \\

    \caption{Impact of varying different training configuration believed to exert implicit regularization pressure when training ResNet18 on CIFAR10 data using SGD \emph{with momentum} (contrast with those without momentum reported in Figure \ref{fig:implicit_regularization} in the main text). Top: varying learning rate. Bottom: varying batch size.}
    \label{fig:llc_curve_implicit_reg_appendix}
\end{figure}

\paragraph{Experiments using SGD \textit{with momentum}}
We repeat the experiments varying learning rate and batch size shown in Figure \ref{fig:implicit_regularization} (top and middle) in the main text, but this time we train ResNet18 on CIFAR10 using SGD \textit{with momentum} instead. The results are shown in Figure \ref{fig:llc_curve_implicit_reg_appendix} and the experimental details are as follow: 
\begin{itemize}
    \item For experiments that varies the learning rate (top), for each learning rate value in $[0.005, 0.05, 0.01, 0.1, 0.2]$ we run SGD with momentum of 0.9 with a fixed batch size of 512 for 30000 iterations. LLC estimations were performed every 1000 iterations with SGLD hyperparameters as follow: step size $\epsilon = 2 \times 10^{-7}$, chain length of 3000 iterations, batch size of 2048 and  $\gamma = 1.0$. 
    \item For experiments that varies the batch size (bottom), for each batch size value in $[16, 32, 64, 128, 256, 512, 1024]$ we run SGD with momentum of 0.9 with a fixed learning rate of 0.01 for 100000 iterations. LLC estimations were performed every 2500 iterations with SGLD hyperparameters as follow: step size $\epsilon = 2 \times 10^{-7}$, chain length of 2500 iterations, batch size of 2048 and  $\gamma = 1.0$. 

\end{itemize}

\paragraph{Explicit $L^2$-regularization}
We also run analogous experiments using explicit $L^2$-regularization. We trained ResNet18 on CIFAR10 dataset using SGD both with and without momentum using the usual cross-entropy loss but with an added $L^2$-regularization term, $\alpha \|w\|^2_2$ where $w$ denote the network weight vector and $\alpha$ is the regularization rate hyperparameter that we vary in this set of experiments. Similar to other experiments in this section, we track LLC estimates over evenly spaced training checkpoints. However, it is worth noting that the loss function used for SGLD sampling as part of the LLC estimation is the original cross-entropy loss function \textit{without} the added $L^2$-regularization term. 

\begin{figure}[h]
    \centering
    \includegraphics[width=0.99\textwidth, keepaspectratio]{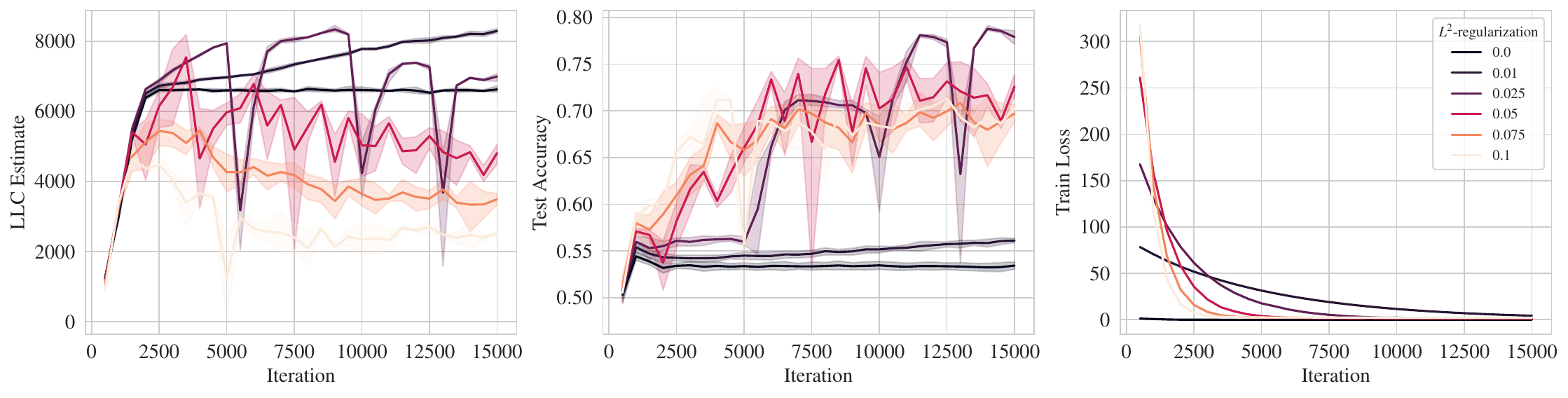}\\
    \includegraphics[width=0.99\textwidth, keepaspectratio]{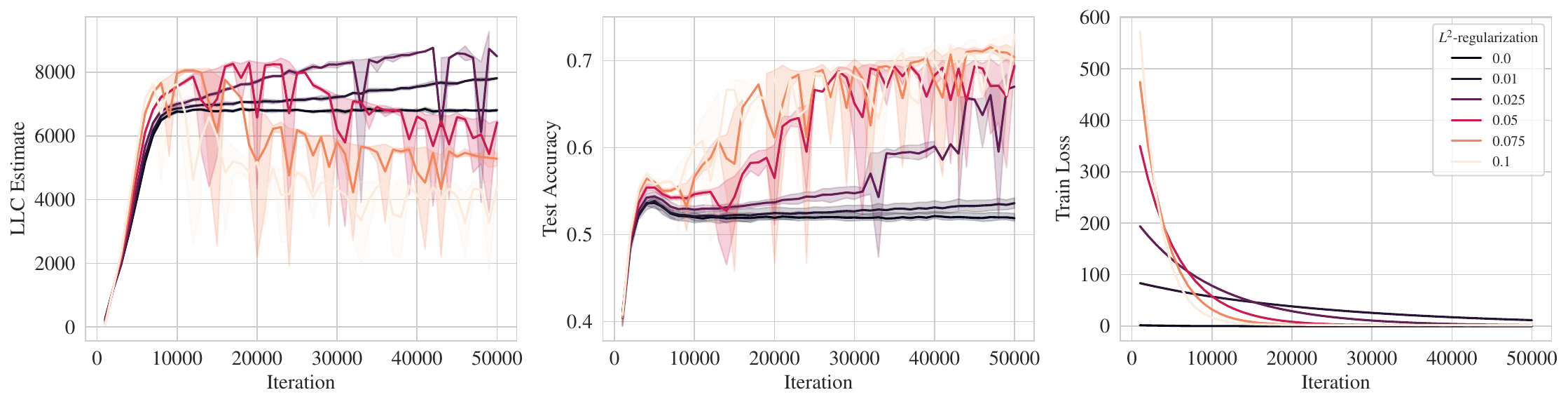} \\
    \caption{Impact of varying explicit $L^2$-regularization rate when training ResNet18 on CIFAR10 data using SGD with (top) and without (bottom) momentum. }
    \label{fig:llc_curve_varyl2reg_appendix}
\end{figure}

The results are shown in Figure \ref{fig:llc_curve_varyl2reg_appendix} and the details are as follow: 
\begin{itemize}
    \item For the experiment \textit{with} momentum (Figure \ref{fig:llc_curve_varyl2reg_appendix} top), for each regularization rate $\alpha \in [0.0, 0.01, 0.025, 0.05, 0.075, 0.1]$, we run SGD with momentum of 0.9 with a fixed learning rate of 0.0005 and batch size of 512 for 15000 iterations. LLC estimations were performed every 500 iterations with SGLD hyperparameters as follow: step size $\epsilon = 5 \times 10^{-8}$, chain length of 2000 iterations, batch size of 2048 and  $\gamma = 1.0$. 
    
    \item For the experiment \textit{without} momentum (Figure \ref{fig:llc_curve_varyl2reg_appendix} bottom), for each regularization rate $\alpha \in [0.0, 0.01, 0.025, 0.05, 0.075, 0.1]$, we run SGD without momentum with a fixed learning rate of 0.001 and batch size of 512 for 50000 iterations. LLC estimations were performed every 1000 iterations with SGLD hyperparameters as follow: step size $\epsilon = 5 \times 10^{-8}$, chain length of 2000 iterations, batch size of 2048 and  $\gamma = 1.0$. 
\end{itemize}

\clearpage

\clearpage
\section{Additional Experiment: LLC for language model}
\label{appendix:expt_language}

\begin{figure}[h]
    \centering
    \includegraphics[width=0.75\textwidth]{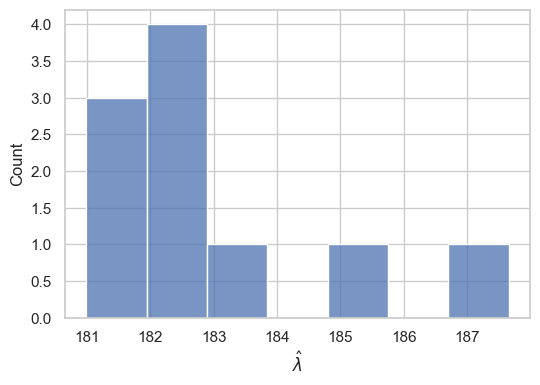}
    \caption{SGLD-based LLC estimates for $\hat{w}^*_n$ at the end of training an attention-only transformer on subset of the Pile dataset. The distribution reports LLC estimates over 10 training repetitions. Again the LLC is a tiny fraction of the 3.3\textbf{m} parameters in the transformer.}
    \label{fig:language}
\end{figure}

The purpose of this experiment is simply to verify that the LLC can be consistently estimated for a transformer trained on language data. In Figure \ref{fig:language}, we show the LLC estimates for $\hat{w}^*_n$ at the end of training over a few training runs. We see the LLC estimates are a small fraction of the 3.3\textbf{m} total parameters. We also notice that the value of the LLC estimates are remarkably stable over multiple training runs. See Appendix \ref{appendix:expt_language} for experimental details.

We trained a two-layer attention-only  (no MLP layers) transformer architecture on a resampled subset of the Pile dataset \cite{gao2020pile,xie2023dsir} with a context length of 1024, a residual stream dimension of $d_{\textrm{model}} = 256$, and 8 attention heads per layer. The architecture also uses a learnable Shortformer embedding \cite{press2021shortformer} and includes layer norm layers. Additionally, we truncated the full GPT-2 tokenizer, which has a vocabulary of around 50,000 tokens, down to the first 5,000 tokens in the vocabulary to reduce the size of the model. The resulting model has a total parameter count of $d=3,355,016$. We instantiated these models using an implementation provided by TransformerLens \cite{nanda2022transformerlens}.

We trained 10 different seeds over a single epoch of 50,000 steps with a minibatch size of 100, resulting in about 5 billion tokens used during training for each model. We used the AdamW optimizer with a weight decay value of $0.05$ and a learning rate of $0.001$ with no scheduler.

We run SGLD-based LLC estimation once at the end of training for each seed at a temperature of $\beta = 1/\log(100)$. We set $\gamma = 100$ and $\epsilon =0.001$. We take samples over 20 SGLD chains with 200 draws per chain using a validation set.

\clearpage
\section{Additional Experiment: MALA versus SGLD}
\label{appendix:expt_mala_vs_sgld}

Here we verify empirically that our SGLD-based LLC estimator (Algorithm \ref{alg:hatlambda_code}) does not suffer from using minibatch loss for both SGLD sampling and LLC calculation. 
Specifically, we compare to LLC estimation via Equation \ref{eq:hatlambda} and the Metropolis-adjusted Langevin Algorithm (MALA), a standard gradient-based MCMC algorithm \cite{roberts1998}. Notice that this comparison rather stacks the odds against the minibatch-version SGLD-based LLC estimator in Algorithm \ref{alg:hatlambda_code} so that it is all the more surprising we see such good results below. 

We test with a two-hidden-layer ReLU network with ten inputs, ten outputs, and twenty neurons per hidden layer. Denote the inputs by $x$, the parameters by $w$, and the output of this network $f(x, w)$. The data are generated to create a ``realizable" data generating process, with ``true parameter" $w^*$: inputs $X$ are generated from a uniform distribution, and labels $Y$ are (noiselessly) generated based on the true network, so that $Y_i = f(X_i, w^*)$. 

Hyperparameters and experimental details are as follows. We sweep dataset size from 100 to 100000, and compare our LLC estimator and the MALA-based one. For all dataset sizes, SGLD batch size was set to 32 and $\gamma = 1.0$, and MALA and SGLD shared the same true parameter $w^*$ (set at random according to a normal distribution). Both MALA and SGLD used a step size of 1e-5 and the asymptotically optimal inverse temperature $\betastar=1/\log n$. Experiments were run on CPU.

\begin{figure}[ht]
\vskip 0.2in
\begin{center}
    \includegraphics[width=\textwidth]{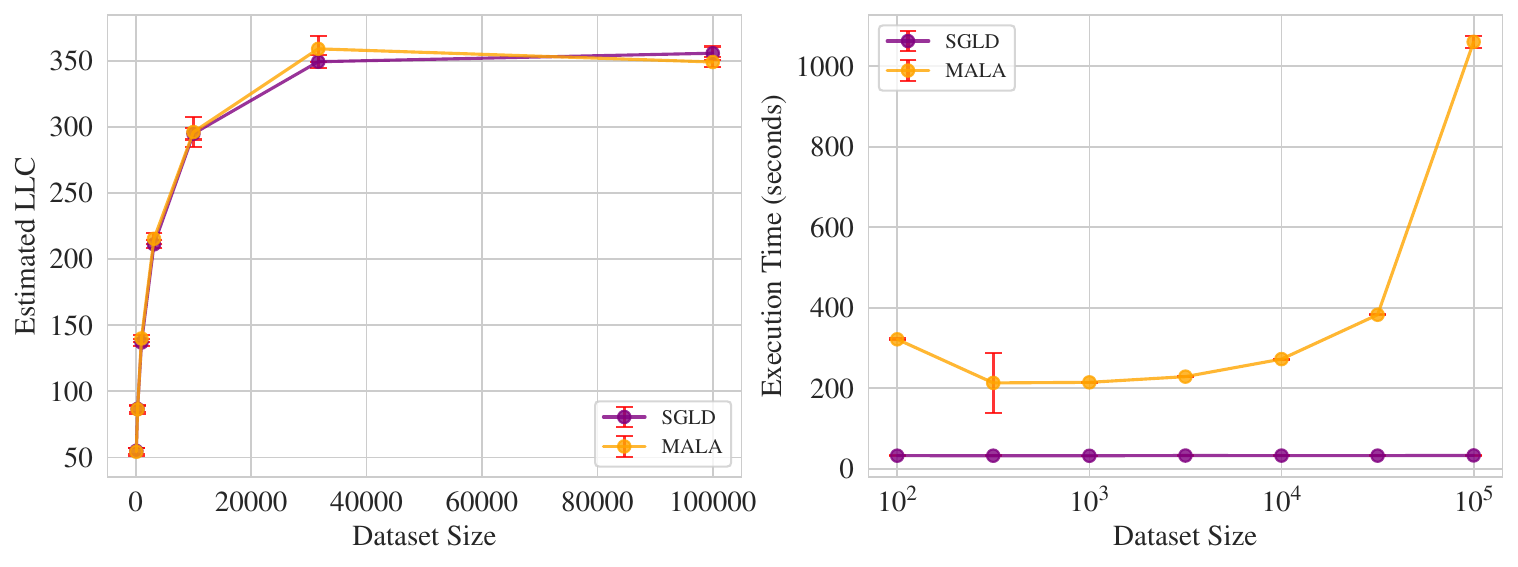}
    \caption{We compare LLC estimation using SGLD with LLC estimation using MALA. They make similar estimates (top) but the SGLD-based method is significantly faster (bottom), especially for large dataset sizes.}
  \label{fig:sgld_vs_mala}
\end{center}
\vskip -0.2in
\end{figure}

The results are summarized in Figure \ref{fig:sgld_vs_mala}. We find that across all dataset sizes, the SGLD and MALA estimates of the LLC agree (Figure \ref{fig:sgld_vs_mala}, top), but the SGLD-based estimate has far lower computational cost, especially as the dataset size grows (Figure \ref{fig:sgld_vs_mala}, bottom).

\clearpage
\section{Additional Experiment: SGD versus eSGD}\label{appendix:expt_esgd}

We fit a two hidden-layer feedforward ReLU network with 1.9m parameters to MNIST using two stochastic optimizers: SGD and entropy-SGD \citep{chaudhariEntropySGDBiasingGradient2019}. We choose entropy-SGD because its objective is to minimize $\proxylocalFn$ over $w^*$, so we expect that the local minima found by entropy-SGD will have lower $\hatlambdawstar$.

\begin{figure}[h]
    \centering
    \includegraphics[width=.75\textwidth]{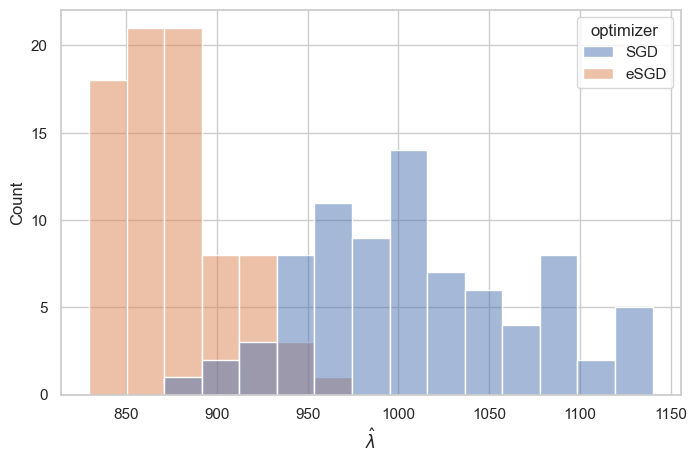}
    \caption{
    LLC estimates for $\hatwstar$ at the end of training a feedforward ReLU network on MNIST. The distribution reports $\hatlambdahatwstar$ over 80 training repetitions where the training data remains fixed in this repetition, only the randomness in the stochastic optimizer is being modded out. We compare two stochastic optimizers -- SGD and entropy-SGD. Note all $\hatlambdahatwstar$ are on the order of $1000$ while the parameter count in the ReLU network is 1.9\textbf{m}. 
    }
    \label{fig:mnist_hatlambda_hist}
\end{figure}

Figure \ref{fig:mnist_hatlambda_hist} shows the LLC estimates $\hatlambdahatwstar$ for $\hatwstar$ at the end of training, optimized by either entropy-SGD or standard SGD. Notice the LLC estimates, for both stochastic optimizers, are on the order of 1000, much lower than the 1.9\textbf{m} number of parameters in the ReLU network.

Figure \ref{fig:mnist_hatlambda_hist} confirms our expectation that entropy-SGD finds local minima with lower LLC, i.e., entropy-SGD is attracted to more degenerate (simpler) critical points than SGD. Interestingly Figure \ref{fig:mnist_hatlambda_hist} also reveals that the LLC estimate has remarkably low variance over the randomness of the stochastic optimizer. Finally it is noteworthy that the LLC estimate of a learned NN model for both stochastic optimizers is, on average, a tiny percentage of the total number of weights in the NN model: $\hatlambdawstar \approx 1000$.

In this experiment, we trained a feedforward ReLU network on the MNIST dataset \cite{deng2012mnist}. The dataset consists of $60000$ training samples and $10000$ testing samples. The network is designed with 2 hidden layers having sizes $[1024, 1024]$, and it contains a total of $1863690$ parameters. 

For training, we employed two different optimizers, SGD and entropy-SGD, minimizing cross-entropy loss. Both optimizers are set to have learning rate of $0.01$, momentum parameter at $0.9$ and batch size of $512$. SGD is trained with Nesterov look-ahead gradient estimator. The number of samples $L$ used by entropy-SGD for local free energy estimation is set to $5$. The network is trained for $200$ epochs. The number of epochs is chosen so that the classification error rate on the training set falls below $10^{-4}$. 

The hyperparameters used for SGLD are as follows: $\epsilon$ is set to $10^{-5}$, chain length to $400$ and the minibatch size 512, and $\gamma=100$. We repeat each SGLD chain $4$ times to compute the variance of estimated quantities and also as a diagnostic tool, a proxy for estimation stability. 


\clearpage
\section{Additional Experiment: scaling invariance}
\label{appendix:expt_rescaling}


In Appendix \ref{appendix:reparam_invariance}, we show theoretically that the theoretical LLC is invariant to local diffeomorphism. Here we empirically verify that our LLC \textit{estimator} (with preconditioning) is capable of satisfying this property in a specific easy-to-test case (though we do not test it in general).

The function implemented by a feedforward ReLU networks is invariant to a type of reparameterization known as \textit{rescaling symmetry}. Invariance of other measures to this symmetry is not trivial or automatic, and other geometric measures like Hessian-based basin broadness have been undermined by their failure to stay invariant to these symmetries \cite{dinh2017}.

For simplicity, suppose we have a two-layer ReLU network, with weights $W_1, W_2$ and biases $b_1, b_2$. Then rescaling symmetry is captured by the following fact, for some arbitrary scalar $\alpha$:
\begin{equation*}
W_2 \text{ReLU}(W_1 x + b_1) + b_2 = \alpha W_2 \text{ReLU}\brac{\tfrac{1}{\alpha} W_1 x + \tfrac{1}{\alpha} b_1} + b_2
\end{equation*}

That is, we may choose new parameters $W_1' = \frac{1}{\alpha} W_1, b_1' = \frac{1}{\alpha} b_1, W_2' = \alpha W_2, b_2' = b_2$ without affecting the input-output behavior of the network in any way. This symmetry generalizes to any two adjacent layers in ReLU networks of arbitrary depth.

Given that these symmetries do not affect the function implemented by the network, and are present globally throughout all of parameter space, it seems like these degrees of freedom are ``superfluous", and should ideally not affect our tools. Importantly, this is the case for the LLC.

We verify empirically that this property also appears to hold for our LLC estimator, \textit{when proper preconditioning is used}.

\subsection{Preconditioned SGLD}

We must slightly modify our SGLD sampler to perform this experiment tractably. This is because applying rescaling symmetry makes the loss landscape significantly anisotropic, forcing prohibitively small step sizes with the original algorithm.

Thus we add preconditioning to the SGLD sampler from Section \ref{section:SGLD_based_wbic}, with the only modification being a fixed preconditioning matrix $A$:

\begin{align}
\label{eq:preconditioned_sgld}
\Delta w_t &= A\frac{\epsilon}{2} \left ( \frac{\beta^* n}{m} \sum_{(x, y) \in B_t} \nabla \log p(y | x, w_t) + \gamma (w^*-w_t) \right )+ N(0, \epsilon)
\end{align}

In the experiment to follow, this preconditioning matrix is hardcoded for convenience, but if this algorithm is to be used in practice, the preconditioning matrix must be learned adaptively using standard methods for adaptive preconditioning \citep{haario1999}.

\subsection{Experiment}

We take a small feedforward ReLU network, and rescale two adjacent layers in the network by $\alpha$ in the fashion described above. We vary the value of $\alpha$ across eight orders of magnitude and measure the estimated LLC using Eq \ref{eq:preconditioned_sgld} for SGLD sampling and Eq \ref{eq:hatlambda} for calculating LLC from samples.\footnote{Note that this means that we are not using Algorithm \ref{alg:hatlambda_code} here, both because of preconditioned SGLD and because we are calculating LLC using full-batch loss instead of mini-batch loss.}

Crucially, we must use preconditioning here so as to avoid prohibitively small step size requirements. In this case, the preconditioning matrix $A$ is set manually, to be a diagonal matrix with entries $\alpha^2$ for parameters corresponding to $W_1$ and $b_1$, entries $\frac{1}{\alpha^2}$ for $W_2$, and entries $1$ otherwise.\footnote{In practical situations, the preconditioning matrix cannot be set manually, and must be learned adaptively. Standard methods for adaptive preconditioning exist in the MCMC literature \cite{haario2001, haario1999}.}

\begin{figure}[!ht]
\vskip 0.2in
\begin{center}
\centerline{\includegraphics[width=\textwidth]{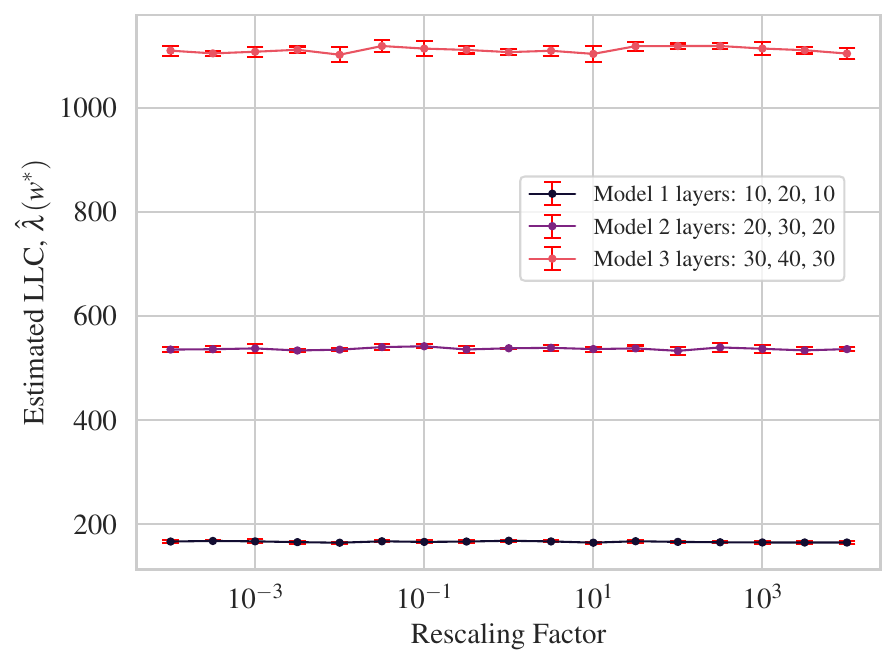}}
\caption{LLC estimation is invariant to rescaling symmetries in ReLU networks. As the rescaling parameter $\alpha$ is varied over eight orders of magnitude, the estimated value of the LLC remains invariant (up to statistical error). The small error bar across multiple SGLD runs illustrates the stability of the estimation method. Model layer sizes, including input dimension is shown in the legend. }
\label{fig:lambdahat-rescaling}
\end{center}
\vskip -0.2in
\end{figure}

The results can be found in Figure \ref{fig:lambdahat-rescaling}. We conclude that LLC estimation appears invariant to ReLU network rescaling symmetries.

\clearpage
\section{Compute resources disclosure} \label{appendix: compute resource disclosure}
Specific details about each type of experiment we carried out are listed below. There addition compute resources required for finding suitable SGLD hyperparameters for LLC estimation, but they did not constitute significant difference to overall resource requirement.

\paragraph{ResNet18 + CIFAR10 experiments. } Each experiment, i.e. training a ResNet18 for the reported number of iteration and performing LLC estimation for the stated number of checkpoints, is run on a node with either a single V100 or A100 NVIDIA GPU depending on availability hosted on internal HPC cluster with 2 CPU cores and 16GB memory allocated. No significant storage required. Each experiment took between a few minutes to 3 hours depending on the configuration (mean: 87.6 minutes, median: 76.7 minutes). 

Estimated total compute is 287 GPU hours spread across 208 experiments.

\paragraph{DLN experiments. } Each experiment, either estimating LLC at the true parameter or at a trained SGD parameter (thus require training), is run on a single A100 NVIDIA GPU node hosted on internal HPC cluster with 2 CPU cores and no more than 8GB memory allocated. No significant storage required. Each experiment took less than 15 minutes. 

Estimated total compute is 150 GPU hours: 600 experiments (including failed ones) each around 15 GPU minutes.

\paragraph{MNIST experiments. } Each repetition of training a feedforward ReLU network using SGD or eSGD optimizer on MNIST data and estimating the LLC at the end of training is run on a single A100 NVIDIA GPU node hosted on internal HPC cluster with 8 CPU cores and no more than 16GB memory allocated. No significant storage required. Each experiment took less than 10 minutes. 

Estimated total compute is 27 GPU hours: 2 sets of 80 repetitions, 10 GPU minutes each.

\paragraph{Language model experiments. }
Training the language model and estimating its LLC at the last checkpoint took around 30 minutes on Google Colab on a single A100 NVIDIA GPU with 84GB memory and 1 CPU allocated. The storage space used for the training data is around 27GB. 

Estimated total compute is 5 GPU hours: 10 repetitions of 30 GPU minutes each.

\end{document}